\documentclass[10pt,twocolumn,letterpaper]{article}

\usepackage{wacv}              % To produce the CAMERA-READY version
\usepackage{graphicx}
\usepackage{amsmath}
\usepackage{amssymb}
\usepackage{booktabs}
\usepackage{multirow}
\usepackage{colortbl}
\usepackage{tabularx}
\usepackage{float}
\usepackage{subcaption}
\usepackage{textcomp}
\usepackage{tikz}
\usepackage[accsupp]{axessibility}

\usepackage[pagebackref,breaklinks,colorlinks]{hyperref}

\usepackage[capitalize]{cleveref}
\crefname{section}{Sec.}{Secs.}
\Crefname{section}{Section}{Sections}
\Crefname{table}{Table}{Tables}
\crefname{table}{Tab.}{Tabs.}

\def\wacvPaperID{1140} % *** Enter the WACV Paper ID here
\def\confName{WACV}
\def\confYear{2025}

\begin{document}

\title{MonoPP: Metric-Scaled Self-Supervised Monocular Depth Estimation by Planar-Parallax Geometry in Automotive Applications}

\author{Gasser Elazab\textsuperscript{1,2} \hspace{10pt} Torben Gr{\"a}ber\textsuperscript{1} \hspace{10pt} Michael Unterreiner\textsuperscript{1} \hspace{10pt} Olaf Hellwich\textsuperscript{2} \\
\hspace{10pt} \textsuperscript{1}CARIAD SE \hspace{10 pt} \textsuperscript{2}Technische Universit{\"a}t Berlin \\
% {\tt\small \{gasser.elazab, michael.unterreiner1\}@cariad.technology}\\
% {\tt\small \{gasser.elazab, olaf.hellwich\}@tu-berlin.de}
}

\maketitle

%%%%%%%%% ABSTRACT
\begin{abstract}
Self-supervised monocular depth estimation (MDE) has gained popularity for obtaining depth predictions directly from videos. However, these methods often produce scale-invariant results, unless additional training signals are provided. Addressing this challenge, we introduce a novel self-supervised metric-scaled MDE model that requires only monocular video data and the camera’s mounting position, both of which are readily available in modern vehicles. Our approach leverages planar-parallax geometry to reconstruct scene structure. The full pipeline consists of three main networks, a multi-frame network, a single-frame network, and a pose network. The multi-frame network processes sequential frames to estimate the structure of the static scene using planar-parallax geometry and the camera mounting position. Based on this reconstruction, it acts as a teacher, distilling knowledge such as scale information, masked drivable area, metric-scale depth for the static scene, and dynamic object mask to the single-frame network. It also aids the pose network in predicting a metric-scaled relative pose between two subsequent images. Our method achieved state-of-the-art results for the driving benchmark \textbf{KITTI} for metric-scaled depth prediction. Notably, it is one of the first methods to produce self-supervised metric-scaled depth prediction for the challenging \textbf{Cityscapes} dataset, demonstrating its effectiveness and versatility. Project page: \href{https://mono-pp.github.io/}{https://mono-pp.github.io/}

\end{abstract}

%%%%%%%%% BODY TEXT
\section{Introduction}
\label{sec:intro}

The process of inferring depth information from a single image, called Monocular Depth Estimation (MDE), is pivotal in the realm of autonomous vehicles and semi-autonomous vehicles. It is crucial for understanding the environment surrounding a vehicle, enabling safe navigation, obstacle detection, and path planning. MDE forms the backbone of perception systems in modern camera-reliant automotive applications, facilitating the development of advanced driver-assistance systems (ADAS)~\cite{chen2015deepdriving}. Further areas of impact include augmented reality~\cite{ConsistentVideoDepth_AR} and robotics~\cite{griffin2020video_robotics}, as summarized by Li \etal in a survey~\cite{MonoDepth_survey}. These use cases leverage MDE to reconstruct and understand the environment without expensive sensors like Lidar or RGB-D cameras~\cite{wang2021research_lidarvscamera}. Chaudhuri~\etal~\cite{chaudhuri1999depth} provide an extensive overview of advances and literature on inferring depth from images. Notably, the revolution of deep learning in MDE has led to significant advancements in regressing depth from single images~\cite{godard2017unsupervised_monodepth1,eigen2014depth,zhou2017unsupervised,liu2015learning,fu2018deep}.

Predicting depth from a single image is inherently an under-constrained problem because a single 2D image lacks sufficient information to uniquely determine the 3D structure of the scene~\cite{MonoDepth_survey}. This ambiguity arises because an infinite number of 3D scenes can project onto the same 2D image, making it challenging to infer absolute depth without additional cues or assumptions~\cite{MonoDepth_survey,godard2017unsupervised_monodepth1}. Therefore, some form of supervision is needed for metric-scaled monocular depth estimation. However, supervised MDE usually requires ground-truth labels from Lidar or RGB-D cameras~\cite{eigen2014depth,geiger2012kitti,ranftl2020towards_zeroshot_crossdata,zhao2024roadbev}, which may not always be available in practical applications. Subsequently, self-supervised MDE has been employed to predict depth at the pixel level using only monocular sequences, stereo pairs, or additional pose information~\cite{godard2017unsupervised_monodepth1}. Most self-supervised methods are trained by novel image synthesis from different views and minimize reconstruction loss via various strategies~\cite{packnet_selfsup,godard2017unsupervised_monodepth1,zhou2017unsupervised,flowdepth2024_selfsup}.

One of the main challenges self-supervised methods encounter is the absence of information about moving objects, which may cause incorrect training signals. This can hinder the network’s optimization and reduce overall quality~\cite{godard2017unsupervised_monodepth1}. One possible solution is to utilize segmentation models to mask out dynamic objects from the scene~\cite{Usingsemanticseg_selfsup_2020}, but these are usually trained with supervision. Godard~\etal introduced an auto-masking loss that mitigates this problem in some scenes~\cite{godard2019digging_monodepth2_selfsup}. In instances where the vehicle’s velocity is available as a reference, Guizilini~\etal~\cite{packnet_selfsup} demonstrate that a scale-aware model can be learned by velocity regression, comparing the traveled distance with the estimated distance. Sui~\etal~\cite{roadaware_SFM_2021_selfsup_scaled} managed to predict scale-aware depth by utilizing camera height.

Watson~\etal~\cite{manydepth} and Guizilini~\etal~\cite{guizilini2022multiframe_toyota} present improvements utilizing multiple frames as input, which significantly increases depth quality. A challenging situation arises when the camera is at a standstill, where the vehicle does not move. Most existing self-supervised methods primarily focus on predicting scale-invariant depth, a characteristic that limits their practical applicability in real-world scenarios, such as automotive applications where accurate, metric-scaled depth information is crucial. Recognizing this gap, our work is specifically oriented towards predicting metric-scaled depth.

In this paper, our focus is on enhancing Monocular Depth Estimation (MDE), which relies solely on a single image as input, thereby eliminating the dependency on camera movement. Concurrently, we train a teacher model to construct the structure using planar-parallax geometry. This model guides the network towards predicting a more reliable and metric-scaled depth, paving the way for practical applications in real-world scenarios. The contributions of this paper can be summarized in three aspects:- 
\begin{itemize}
    \item Through comparative analysis with other self-supervised metric-scaled monocular methods, our method demonstrated superior performance on the KITTI dataset. Moreover, our method is among the first to achieve self-supervised metric-scaled depth on the challenging Cityscapes dataset.

    \item By utilizing a single piece of information --- the camera’s mounting location above the ground --- we have shown that a self-supervised MDE is capable of producing reliable metric depth results, making it highly applicable in vehicular perception tasks.

    \item To the best of our knowledge, this is the first model to leverage planar-parallax geometry for self-supervised metric-scaled scene reconstruction without the need for any ground-truth information
\end{itemize}
%-------------------------------------------------------------------------
\section{Related Work}
\label{sec:related_work}
The field of monocular depth estimation has witnessed major advancements~\cite{MonoDepth_survey}, especially since the deep learning breakthrough by AlexNet~\cite{alexnet_history_survey,alexnet_org}. This section covers the evolution from general monocular depth estimation to self-supervision, the history of Planar-Parallax geometry, and recent efforts in predicting scale-aware self-supervised monocular depth.

\textbf{Monocular Depth Estimation}. The real breakthrough came with the application of Convolutional Neural Networks (CNNs) between 2016 and 2018~\cite{li2021survey_CNN}. One of the pioneering models in this domain is MiDAS~\cite{Midas_2022_main}, it employs an encoder-decoder architecture, where the encoder extracts high-level features, and the decoder generates the depth map through up-sampling techniques. MiDAS is trained on multiple datasets, enabling robust performance across diverse conditions and environments. They have also utilized vision transformers that enhanced generalizability and accuracy in MiDaS V3.0~\cite{birkl2023midas_MIDASV3}. In addition, Depth-Any-thing~\cite{depthanything} demonstrated generalizability across datasets by utilizing a hybrid encoder that integrates CNNs with Vision Transformers, along with an attention mechanism to focus on the relevant part of the image, thereby improving the accuracy of depth estimation. Despite these models’ ability to generate relative depth information accurately, they still require some scale information or fine-tuning to predict accurately scaled depth values.

% \subsection{}
% \label{subsec:self-supervised-mono-depth}
\textbf{Self-Supervised Monocular Depth Estimation}. There has been significant advancement in the fully self-supervised MDE utilizing analogous encoder-decoder architecture. For example, Godard \etal~\cite{godard2017unsupervised_monodepth1,godard2019digging_monodepth2_selfsup} introduced two models, marking the beginning of the era of self-supervised monocular depth estimation. Their work introduced innovative techniques that leveraged stereo image pairs and monocular video sequences to train depth estimation models without requiring any ground-truth information, significantly advancing the field and inspiring numerous subsequent research efforts~\cite{litemono_2023_selfsup,Usingsemanticseg_selfsup_2020}.

LiteMono~\cite{litemono_2023_selfsup} and VADepth~\cite{VADEPTH} utilized combinations of CNNs and attention blocks, enhancing the capture of multi-scaled local features and long-range global context. Although LiteMono~\cite{litemono_2023_selfsup} is a lightweight model suitable for real-time use, it also outperforms Monodepth2~\cite{godard2019digging_monodepth2_selfsup}. Moreover, there are emerging methods such as manydepth~\cite{manydepth} and depthformer~\cite{guizilini2022multiframe_toyota} that take advantage of multiple images at test time, they have achieved superior results when multiple frames are available. However, multi-frame methods do not predict the same depth quality in single-frame scenarios. Most self-supervised MDEs use the per-frame median ground truth to scale predictions, allowing for comparison of relative depth values with the ground truth. Directly predicting metric-scaled depth from self-supervision is inherently ill-posed, necessitating the calculation of true scale using prior knowledge about the actual environment.

\textbf{Metric-Scaled Self-Supervised MDE}. PackNet-SfM~\cite{packnet_selfsup} is a novel deep learning architecture that leverages 3D packing and unpacking blocks to effectively capture fine details in monocular depth map predictions. In PackNet-SfM~\cite{packnet_selfsup}, they also managed to achieve scale-aware monocular depth results by using only the velocity signal of the vehicle, similar to the model developed by Tian~\etal~\cite{ScaleRecoveryOdometry_2021} for predicting metric-scaled odometry from sequence of images. Additionally, Wagstaff~\etal~\cite{wagstaff_scalerecovery_2021_SELFSUP} demonstrated that metric-scaled depth can be predicted from a moving monocular camera by utilizing only the camera’s mounting position above the ground. Their study reported favorable results when pose ground-truth information was accessible. However, the results were less satisfactory when the scale was inferred solely from the camera’s height. Besides, Sui \etal~\cite{roadaware_SFM_2021_selfsup_scaled} utilized ground-truth camera height information to retrieve the true scale of predictions by aligning the planar road. However, the accuracy of these results was inferior compared to those obtained using pose information. In addition, we have examined a concurrent work, which is still published as a pre-print~\cite{CameraHeightDoesnChange}, this study utilizes large-scale data scraped from the internet to establish prior object sizes and relies on the long temporal dependency of sequences to calculate metric-scale depth.

\textbf{Planar-Parallax Geometry}. Planar-parallax geometry, introduced in~\cite{PP_foundation_94_1,sawhney19943d}, derives a 3D structure relative to a planar surface. It decomposes changes in the static scene into planar homography and parallax residual epipolar flow. This flow correlates with the structure of the non-dynamic scene across multiple frames, improving reconstruction accuracy and stability~\cite{sawhney19943d,MRFLow}. Recently, Xing \etal~\cite{xing2022joint_PP_HAO_TUM} also used the output of planar parallax geometry as a complementary training signal along with the ground-truth to predict accurate depth values. Moreover, recent trainable methods utilize two planar-aligned frames at once to detect depth, achieving better accuracy than their predecessor methods~\cite{RoadPlanarParallax,PP_jointPrediction_XING,xing2022joint_PP_HAO_TUM}.

% One of the benefits is that this residual flow establishes a straightforward mathematical correlation between the structure of a stationary scene and the residual flow across multiple-view frames, given that they are aligned with respect to a certain plane.
\section{Method}
\label{sec:method}

In this section, we present our design motivation for our approach. We employ a multi-frame teacher network grounded in planar-parallax geometry, enabling the joint optimization of metric-scaled depth and pose. This teacher network distills knowledge to a single-frame model to predict accurate depth information. In addition, by confining the teacher’s output to reconstruct only static scenes, we simultaneously generate a static mask to facilitate the training of single-frame model. Unlike recent studies~\cite{godard2017unsupervised_monodepth1,godard2019digging_monodepth2_selfsup,CADepth_2021}, we did not use fixed data-specific depth binning for converting network output to interpretable depth. Instead, we rely on residual flow binning, which is intrinsically tied to the scene structure, ensuring depth range adaptability per frame. This approach leverages the benefits of planar-parallax geometry. On the other hand, most existing methods that utilize cost volumes, transformers, or similar architectures are computationally intensive. As a result, they are not suitable for real-time performance without significant computational resources~\cite{litemono_2023_selfsup}. To address this issue, we use a computationally expensive model to distill knowledge into a more efficient deployment model.

In planar-parallax geometry~\cite{irani1996parallax}, metric-scaled values can be obtained relative to a correctly aligned planar scene. However, similar to Structure from Motion (SfM)~\cite{zhou2017unsupervised}, planar-parallax geometry requires a baseline, which may not always be available in vehicular applications, such as stationary scenarios or minimal baselines in traffic jams. To address this, we implemented the planar-parallax module as a teacher, that predicts the static structure details, correctly scales the pose, and contributes to masking dynamic objects. This approach ensures our model’s robustness and adaptability to various scenarios.

Our method comprises three main pipelines. The first pipeline uses a single image to generate a disparity map, which is then converted to depth maps. The second, the Planar-Parallax pipeline, warps neighboring planar-aligned views to the target view, outputs residual flow, and then computes the scene’s structure and depth. Predicting residual flow directly simplifies the problem by transforming it into a 1D epipolar flow matching problem.

\cref{subsec: planar-parallax-form} illustrates the mathematical formulation of the Planar Parallax, detailing how the residual flow between a planar-aligned source frame and a target frame is correlated with the structure of the point in 3D space. Then,~\cref{subsec:network} covers the Planar Parallax pipeline, including its components. In addition, it presents the monocular depth estimation network and the pose network. Finally,~\cref{subsec:training_losses} explains the formulation of the loss functions and the training strategy used to train the entire pipeline.

\subsection{Planar-Parallax formulation}
\label{subsec: planar-parallax-form}

\begin{figure}[h]
  \centering
   \includegraphics[width = 1.0 \linewidth,height=7.5 cm]{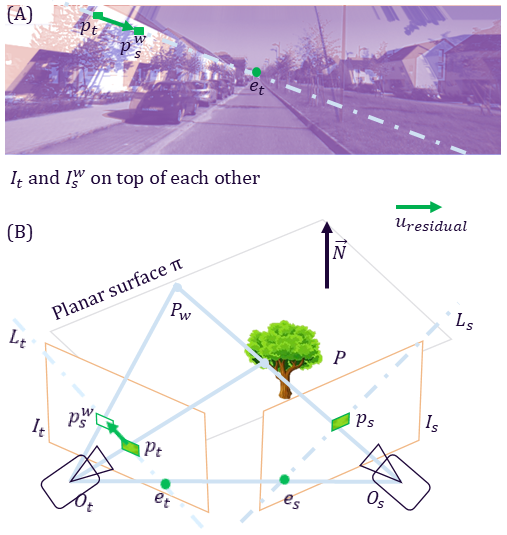}

   \caption{ (A) Example from two sequential frames of KITTI, aligned by the planar road homography, it is obvious that residual flow increases as the height of the object increases relative to the road. (B) Illustrative example of the epipolarity of the residual flow between $p_t$ and $p^w_s$, the figure is inspired by~\cite{RoadPlanarParallax}}
   \label{fig:warping_basic}
\end{figure}

The foundations of Planar-parallax geometry are in~\cite{irani1996parallax,RoadPlanarParallax,PP_jointPrediction_XING}. In essence, given two views of a scene, the goal is to align the source frame $I_s$ to the target frame $I_t$ with respect to a planar surface $\pi$, as shown in~\cref{fig:warping_basic}. This alignment uses a planar homography $H_{s \rightarrow t}$, mapping points from the source image $p_s$ to the target image $p_t$.

By applying $H_{s \rightarrow t}$, each point $p_s$ in the source image is transformed to a new position $p_{s}^{w}$ in the warped source image $I_{s}^{w}$. $H_{s \rightarrow t}$ is calculated by the plane normal and the relative pose between $I_t$ and $I_s$, as shown in~\cref{eqn:homography_calc}:

\begin{equation} \label{eqn:homography_calc} H_{s \rightarrow t} = K (R_{s \rightarrow t} + \frac{\vec{T_{s \rightarrow t}}\vec{N^T}}{h_c}) K^{-1} \end{equation}

In~\cref{eqn:homography_calc}, $K$ is the camera's intrinsic matrix, $R_{s \rightarrow t}$ and $T_{s \rightarrow t}$ are the rotation and translation matrices from the source view $O_{s}$ to the target view $O_{t}$, $\vec{N^T}$ is the normal vector of the planar surface $\pi$, and $h_c$ is the distance between $O_{t}$ and $\pi$, typically the camera height from the ground in vehicular applications.

After alignment, static objects exhibit parallax residual flow, described in~\cref{eqn:residual_flow}, which provides insights into the relative motion, depth, and height of objects relative to the planar surface. This flow is based on differences in apparent motion between planar-aligned points of $I_s$ and $I_t$, offering a framework to understand and manipulate visual depth cues~\cite{irani1996parallax}.

\begin{equation}
\label{eqn:residual_flow}
    u_{s\rightarrow t}^{\text{res}} = \frac{-\gamma \cdot \frac{T_z}{h_c}}{1 - \gamma \cdot \frac{T_z}{h_c}} \cdot (p_t - e_t)
\end{equation}

$u_{s\rightarrow t}^{\text{res}}$ is the residual parallax between the point of the warped image $p_s^w$ and its correspondent in the target image $p_t$. $\gamma_{p}$ is the structure of this point equating to $\frac{h_p}{d_p}$, where $h_p$ is the height of this point from the planar surface $\pi$ and $d_p$ is the depth of this point from the target frame. $e_t$ is the epipole of the target frame. $T_z$ is the translation in z-direction. This equation is valid only if $T_z \neq 0 $. Full proofs and derivations of the equations involved in calculating the parallax residual flow can be found in~\cite{sawhney19943d,irani1996parallax}

\subsection{Problem setup}
\label{subsec:network}

\begin{figure*}
  \centering

  \hfill

    % \fbox{\rule{0pt}{2in} \rule{.9\linewidth}{0pt}}
    \includegraphics[width=1.0\textwidth]{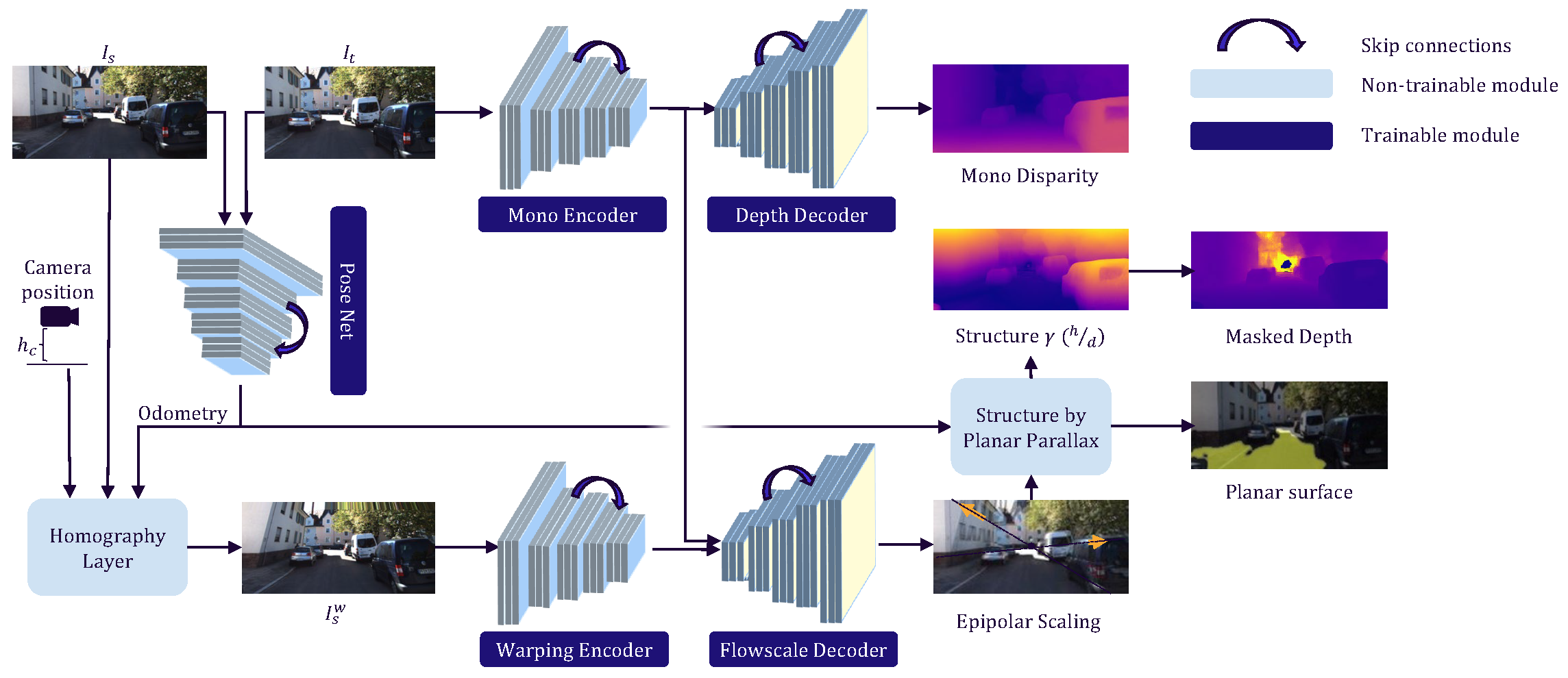}
    \caption{Our framework is composed of two primary pipelines. The first pipeline performs monocular depth estimation by using a single image, $I_t$, as input. The second pipeline aims to reconstruct the geometry by determining the scale for the previously warped image, $I_{s}^{w}$. It then calculates the structure from this information and serves a dual purpose: it distills information to the monocular depth estimator to learn reliable depth about the static scenes, and it provides a mask to filter out dynamic objects. Regarding the colormap, brighter yellow means higher values, and vice versa. All images are cropped by the same ratio for better visualization.}
  \label{fig:full_pipeline}
\end{figure*}

The network consists of two pipelines: one is the monocular depth estimation network used at test time, and the other is the planar-parallax pipeline, utilized as a teacher module and used only during training. Given two sequential images $I_{t}$ and $I_{s}$, they are fed to a pose network which is a ResNet-18~\cite{ResNet_oirignal}, similar to~\cite{godard2019digging_monodepth2_selfsup}. This network produces the relative rotation and translation between the two frames, which are then used for computing losses, computing homography, and calculating the depth from Planar-Parallax, as shown in~\cref{fig:full_pipeline}. 

The homography is computed from the relative pose information and the camera height prior to the planar surface, based on~\cref{eqn:homography_calc}. Then $I_{s}$ is warped by the computed homography, and then fed to a warping encoder which is a ResNet-50~\cite{ResNet_oirignal}, pretrained on ImageNet~\cite{russakovsky2015imagenet}. Another similar encoder, the 'Mono encoder', is also utilized to encode the target image $I_{t}$, this is the encoder used at test time. The outputs of the two encoders are concatenated and passed to the Flowscale decoder, a series of CNNs for upsampling the input and predicting the output for multi-scale stages, such as~\cite{godard2019digging_monodepth2_selfsup,manydepth}. The depth decoder has a similar architecture to the Flowscale decoder, but only takes the encoded information of $I_t$ as an input, and this is the one used at test time as well, as shown in~\cref{eqn:networks_mono}.

\begin{equation}
\label{eqn:networks_mono}
    D_{t}^{\text{mono}} = \theta_{\text{mono}}(I_{t})
\end{equation}

There are two main trainable networks, which are $\theta_{\text{mono}}$ and $\theta_{\text{pp}}$, as shown in~\cref{eqn:networks_mono,eqn:networks_pp}. $D^{\text{mono}}_{t}$ is the calculated depth from the disparity produced by $\theta_{\text{mono}}$.

% \begin{equation}
% \label{eqn:networks_mono}
%     D_{t}^{\text{mono}} = \theta_{\text{mono}}(I_{t}) \quad \text{and} \quad s_{t} = \theta_{\text{pp}}(I_{t}, I_{s}^{w})
% \end{equation}

\begin{equation}
\label{eqn:networks_pp}
     s_{t} = \theta_{\text{pp}}(I_{t}, I_{s}^{w})
\end{equation}

 The output of $\theta_{\text{pp}}$ is a single value per pixel, which is a \textit{scaling} value representing the scaling quantity to be multiplied by the $(p_t - e_t)$ in equation~\cref{eqn:residual_flow}. The output of the Flowscale decoder is mapped to specific bins which are adjusted for different image resolutions; hence the scaled $s_{t}$ is transformed to $S_{t}$. Then, these values are multiplied by $(p_t - e_t)$ to calculate the $u_{s\rightarrow t}^{\text{res}}$, as shown in~\cref{eqn:calc_gamma}. 
\begin{equation}
    \label{eqn:calc_gamma}
    S_{t} = \frac{u_{s\rightarrow t}^{\text{res}}}{q_{t}-e_{t}}
\end{equation}

After $u_{s\rightarrow t}^{\text{res}}$ is calculated from $S_{t}$, we can calculate $\gamma$ for this point given $T_z$ and $h_c$, as shown in~\cref{eqn:calc_gamma,eqn:calc_gamma_2}. $T_z$ is derived directly from the output of the PoseNet, while $h_c$ is the camera's height from the ground.

\begin{equation}
    \label{eqn:calc_gamma_2}
    \gamma = \frac{S_{t} }{S_{t}+1} \cdot \frac{h_c}{T_{z}}
\end{equation}
Such that the $S_{t}$ is the scaled output of $\theta_{\text{pp}}$ representing the offset of the pixel along the epipolar line. This is how $\gamma$ is retrieved from the network's output. Then the depth or height is derived from $\gamma$ based on~\cref{eqn:depth_from_gamma}, the derivation of the equation is in~\cite{RoadPlanarParallax}.
\begin{equation}
    \label{eqn:depth_from_gamma}
    D_{t}^{\text{pp}} = \frac{h_{c}}{\gamma + \vec{N^{T}} \cdot (K^{-1}p)}
\end{equation}

The required outputs of this pipeline to calculate the losses are $D_{t}^{\text{mono}}$, $u_{t}^{\text{res}}$, and $D_{t}^{\text{pp}}$. These can synthesize novel images from the neighboring views, as shown in~\cref{eqn:syntheize_new_view_depth}, to compute the photometric reprojection losses. 
\begin{equation}
    \label{eqn:syntheize_new_view_depth}
    \hat{I_{t}^{d}} = I_{s} \langle \text{proj}(D_{t},R_{t \rightarrow s},T_{t \rightarrow s})\rangle
\end{equation}

where $\text{proj}()$ is the resulting coordinates of the projected depths, and $\langle\rangle$ is the bi-linear sampling operator, which is locally sub-differentiable~\cite{billinear_sampling}. However, novel views can also be constructed by $u^{\text{res}}_{s\rightarrow t}$, as shown in~\cref{eqn:syntheize_new_view_flow}.

\begin{equation}
    \label{eqn:syntheize_new_view_flow}
    \hat{I_{t}}^{\text{res}} = I_{s}^{w} \langle u_{t}^{\text{res}}\rangle
\end{equation}

One drawback of the non-learnable calculation of depth from the residual flow via planar-parallax geometry is the noise around the epipole, as $p_{t}-e_{t}$ will be minimal and the parallax is negligible, which may disturb the learning process and produce incorrect depths. This is why we proposed a novel certainty mask for this case, which we call \textit{certainty mask}. It masks out the pixels resulting from any errors from the $u^{\text{res}}_{ s\rightarrow t}$ around the epipole, by only including the pixels, which are reconstructed by the $D_{t}^{pp}$ to the same one reconstructed by $u^{\text{res}}_{ s\rightarrow t}$, as presented in~\cref{eqn:certainity_mask}.

\begin{equation}
    \label{eqn:certainity_mask}
    M_{\text{cert}} = \begin{cases} 
    1 & \text{if } |\langle u_{s\rightarrow t}^{\text{res}}\rangle - \langle \text{proj}(D_{t}^{\text{pp}},R_{t \rightarrow s},T_{t \rightarrow s})\rangle| \leq \epsilon\\
    0 &\text{otherwise}
    \end{cases}
\end{equation}

To measure the correctness of the computed homography, as in~\cref{eqn:homography_calc}, the planar road surface must be detected. For this, a flat area detector, inspired by~\cite{yang2018lego}, is introduced. The flat area is detected by calculating the surface normal vectors of each pixel based on its neighboring points. For each pixel $p_{i}$, we compute the normal vector by cross product of the vector of $p_{i}$ with its neighbors. First, $p_{i}$ is projected in the three-dimensional space by the predicted depth $D_{i}$, such that $P_{i} = D_{i}\cdot K^{-1} p_{i}$. Then, the average normal vector is computed using cross products of neighboring vectors. This average normal vector is used to calculate the planar area mask, as shown in~\cref{eqn:final_surface_normal}. More details are explained in the supplementary materials.

\begin{equation}
    \label{eqn:final_surface_normal}
    M_{flat} = \begin{cases} 
    1 & \text{if } | \text{cos\_sim}(\frac{\mathbf{N}_{\text{avg}}(i, j)}{\|\mathbf{N}_{\text{avg}}(i, j)\|},\vec{N^T})| > \tau \\
    0 &\text{otherwise}
    \end{cases}
\end{equation}

In~\cref{eqn:final_surface_normal}, $\text{cos\_sim}$ is the cosine similarity between vectors, and $\tau$ is the threshold for determining whether it is a flat surface. To focus on the relevant region of the image that represents the road, a trapezoidal mask centered in the image isolates the road by including only central pixels with a structure near zero, $|\gamma|\leq 0.05$. This mask selectively includes only the road plane from other flat regions, and this is our utilized flat surface detector in~\cref{fig:full_pipeline}. 

\subsection{Training}
\label{subsec:training_losses}

In our approach, we have five main losses, which are $L_{\text{homo}}$, $L_{\text{mono}}$, $L_{\text{pp}}$, $L_{\text{cons}}$, and $L_{\text{res}}$. First, the photometric reprojection loss utilized within some losses is L1 and SSIM~\cref{eqn:photometric_loss_reproj}, as introduced in~\cite{godard2019digging_monodepth2_selfsup,manydepth,godard2017unsupervised_monodepth1,reproj_loss}.

\begin{equation}
    \label{eqn:photometric_loss_reproj}
    \text{pe}(I_{a},I_{b}) = \frac{\alpha}{2}(1-\text{SSIM}(I_{a},I_{b})) + (1-\alpha) \lVert I_{a} - I_{b} \lVert
\end{equation}

Homography loss is responsible for accurate road-planar homography estimation, and it is also the one encouraging the pose's output to be metric-scaled. It encourages the planar area of $I^{w}_{s}$ to be similar to the planar area of $I_{t}$. Hence, it enforces the correct alignment of the road planar surface, as shown in~\cref{eqn:LOSS_HOMO}.
\begin{equation}
    \label{eqn:LOSS_HOMO}
    L_{\text{homo}} = \frac{1}{\sum M_{\text{flat}}}\sum_{p_{i}}^{N} M_{\text{flat}} (\text{pe}(I_{s}^{w},I_{t}))
\end{equation}
\begin{table*}[!ht]
\centering
\footnotesize % This line will decrease the font size
\setlength\tabcolsep{4.8 pt}
\renewcommand{\arraystretch}{1.0}
\begin{tabular*}{\textwidth}{@{\extracolsep{\fill}}|c|c|c|c|c|c|c|c|c|c|c|}
\hline
 & Year & Method & Train & Abs Rel $\downarrow$ & Sq Rel $\downarrow$ & RMSE $\downarrow$ & RMSE log $\downarrow$ & $\delta < 1.25$ $\uparrow$ & $\delta <1.25^2$ $\uparrow$ &  $\delta <1.25^3$ $\uparrow$ \\
 \cline{2-11}    
  \multirow{10}{*}{\rotatebox[origin=c]{90}{scaled by GT}} & 2019 & {Monodepth2~\cite{godard2019digging_monodepth2_selfsup}}  & M & 0.115 & 0.903 & 4.863 & 0.193 & 0.877 & 0.959 & 0.981 \\
 \cline{2-11}  

 & 2020 & PackNet-SFM~\cite{packnet_selfsup}  &  M & 0.111 & 0.785 & 4.601 & 0.189 & 0.878 & 0.960 & 0.982 \\
 \cline{2-11}  

 % & 2020 & DNet~\cite{DNet}  &  M & 0.113 & 0.864 & 4.812 & 0.191 & 0.877 & 0.960 & 0.981 \\
 % \cline{2-11}  

 % & 2021 & ManyDepth~\cite{manydepth} & M & 0.106 & 0.818 & 4.750 & 0.196 & 0.874 & 0.957 & 0.979 \\
 %  \cline{2-11}  

 & 2021 & CADepth~\cite{CADepth_2021} &  M & 0.110 & 0.812 & 4.686 & 0.187 & 0.882 & 0.961 & 0.981 \\
 \cline{2-11}  

 & 2022 & VADepth~\cite{VADEPTH} &  M & 0.104  &0.774 &4.552 & 0.181 & 0.892  &0.965 & 0.983 \\
 \cline{2-11}  
 & 2022 & MonoFormer~\cite{monoformer}  &  M & 0.108  &0.806 &4.594 & 0.184 & 0.884  &0.963 & 0.983 \\
 \cline{2-11}  
 \cline{2-11}  
 & 2022 & MonoViT~\cite{zhao2022monovit}  &  M & \underline{0.099} & \underline{0.708}  & \underline{4.372} & \underline{0.175} & \underline{0.900}  &\underline{0.967} & \underline{0.984} \\
 \cline{2-11}  

 & 2023 & Lite-Mono~\cite{litemono_2023_selfsup}  &  M & 0.107  &0.765 &4.561 & 0.183 & 0.886  &0.963 & \underline{0.983} \\
 \cline{2-11}  

 & 2023 & Lite-Mono-S~\cite{litemono_2023_selfsup}  &  M & 0.110  &0.802 &4.671 & 0.186 & 0.879  &0.961 &0.982 \\
 \cline{2-11}  
  & 2023 & TriDepth~\cite{tridepth}  &  M & \textbf{0.093}  & \textbf{0.665} & \textbf{4.272} & \textbf{0.172} & \textbf{0.907}  & \textbf{0.967} &\textbf{0.984} \\
 \cline{2-11}  
 
 &  \multicolumn{2}{c|}{MonoPP (ours) }  &  M & 0.105  &0.776 &4.640 & 0.185 & 0.891   &0.962 &0.982 \\
\hline
\hline

 \multirow{11}{*}{\rotatebox[origin=c]{90}{w/o scaling}} & 2019 & Monodepth2**~\cite{godard2019digging_monodepth2_selfsup}  &  M+camH &   0.126  &   0.973  &   4.880  &   0.198  &   0.864  &   0.957  &   0.980  \\
\cline{2-11}
 & 2020 & DNet~\cite{DNet}  &  M+camH & 0.118  &0.925 &  4.918  & 0.199 &0.862 &0.953  &0.979 \\
 \cline{2-11}  
 & 2020 & Zhao~\etal ~\cite{depth_by_GANS} &  M+SC &  0.146  &1.084 &5.445 & 0.221 & 0.807  &0.936 &0.976\\
 \cline{2-11}  
 & 2020 & PackNet~\cite{packnet_selfsup}  &  M+V &  0.111  &0.829 &4.788 & 0.199 & 0.864  &0.954 &0.980\\
 \cline{2-11}  
 & 2021 & Wagstaff~\etal ~\cite{wagstaff_scalerecovery_2021_SELFSUP}  &  M+Pose &  0.123  &0.996 &5.253 & 0.213 & 0.840  &0.947 &0.978 \\
 \cline{2-11}  
 & 2021 & Wagstaff~\etal ~\cite{wagstaff_scalerecovery_2021_SELFSUP} &  M+camH &  0.155  &1.657 &5.615 & 0.236 & 0.809  &0.924 &0.959 \\
 \cline{2-11}  

& 2021 & Sui~\etal~\cite{roadaware_SFM_2021_selfsup_scaled}  &  M+camH & 0.128  &0.936 &  5.063  & 0.214 & 0.847 &0.951  &0.978 \\
 \cline{2-11}  
 & 2022 & VADepth~\cite{VADEPTH}  &  M+camH &  0.109  & \textbf{0.785} & \textbf{4.624} & 0.190 & 0.875  & \underline{0.960} & \textbf{0.982} \\
 \cline{2-11}  
  & 2022 & DynaDepth~\cite{dynadepth} &  M+Pose &  0.109  & \underline{0.787} & 4.705 & \underline{0.195} & {0.869}  & {0.958} & \underline{0.981} \\
        
 \cline{2-11}  
 & 2023 & Lee~\etal~\cite{Scaleaware_visualInertial_selfsup_scaled}  &  M+Pose &  0.141 &1.117 &5.435 & 0.223 & 0.804  &0.942 &0.977 \\
 \cline{2-11}  
 & 2024 & FUMET~\cite{CameraHeightDoesnChange}  &  M+SI &  \underline{0.108} &\textbf{0.785} &4.736 & 0.195 & \underline{0.871}  &0.958 &\underline{0.981} \\
 \cline{2-11}
 
&  \multicolumn{2}{c|}{MonoPP (ours) }  &  M+camH &   \textbf{0.107}  &   0.835 &   \underline{4.658}  &   \textbf{0.186}  &   \textbf{0.891}  &  \textbf{0.962}  &   \textbf{0.982}   \\   
\hline

\end{tabular*}
\caption{Comparison of our method to existing self-supervised approaches on the KITTI~\cite{geiger2012kitti} Eigen split~\cite{eigen2015predicting}. This comparison only includes the single-frame methods, the more inclusive study is provided in the supplementary materials. As shown, there are two separate tables. The upper one is dedicated for the comparison of scale-invariant depth, which means the predicted depth is scaled per-frame with the median of the ground-truth (GT), $D_{\text{scaled}}= \frac{\text{med}(D_{\text{gt}})}{\text{med}(D_{\text{pred}})} \cdot D_{\text{pred}}$. The lower table focuses on comparing against the methods that predict scaled depth. The best results in each subsection are in \textbf{bold}, and the second-best are \underline{underlined}. All comparisons are done for the medium resolution (640 x 192). \textbf{M} stands for training by monocular videos, and \textbf{S} includes stereo data as well. \textbf{SC*} stands for predicting a scale-consistent output, which may still need GT for scaling. \textbf{Pose} for utilizing the pose information, \textbf{V} for utilizing the vehicle's velocity, \textbf{camH} for utilizing camera height from the ground, and \textbf{SI} for scraping large datasets from the internet during training. $\uparrow$ higher values are better. $\downarrow$ lower values are better. ** a baseline that we implemented to predict post-processed metric-scaled depth from Monodepth2, scaled by the GT camera height.}

\label{tab:kitti_tab}
\end{table*}

where $\text{pe}$ is the reprojection loss, illustrated in~\cref{eqn:photometric_loss_reproj}, and $N$ is the number of pixels. This loss is calculated for all available source images and for all multi-scale outputs, similar to~\cite{godard2019digging_monodepth2_selfsup,manydepth}.

$L_{\text{mono}}$ and $L_{\text{pp}}$ similarly minimize the reprojection of the predicted depth, whether by the monocular depth estimation or the planar-parallax module. In addition, auto-masking $M_{\text{auto}}$ is implicitly utilized, as explained in~\cite{godard2019digging_monodepth2_selfsup}.

\begin{equation}
    \label{eqn:LOSS_mono}
    L_{\text{mono}} = \frac{1}{\sum M_{\text{auto}}}\sum_{p_{i}}^{N} M_{\text{auto}}(\text{pe}(\hat{I}_{t}^{d_{\text{mono}}},I_{t}))
\end{equation}

However, one difference is that the depth by planar-parallax is masked by the certainty mask, as shown in~\cref{eqn:LOSS_pp}.

\begin{equation}
    \label{eqn:LOSS_pp}
    L_{\text{pp}} = \frac{1}{\sum M_{\text{cert}} \cdot M_{\text{auto}}}\sum_{p_{i}}^{N} M_{\text{auto}} \cdot M_{\text{cert}} (pe(\hat{I}_{t}^{d_{\text{pp}}},I_{t}))
\end{equation}

The residual loss quantifies computing the error between $\hat{I_{t}}^{\text{res}}$ and $I_{t}$, which ensures that the scaling output from $\theta_{pp}$ represents the correct epipolar scaling, which is residual flow, as shown in~\cref{eqn:LOSS_res}.
\begin{equation}
    \label{eqn:LOSS_res}
     L_{\text{res}} = \frac{1}{N}\sum_{p_{i}}^{N} (\text{pe}(\hat{I}_{t}^{\text{res}},I_{t}))
\end{equation}

The consistency mask, presented in~\cref{eqn:matching_mask} and inspired by~\cite{manydepth}, checks the pixels where the depth $D_{t}^{\text{mono}}$ and $D_{t}^{\text{pp}}$ agree to a certain threshold $\delta$.
\begin{equation}
    \label{eqn:matching_mask}
    M_{\text{static}} = \text{max}( \frac{D_{t}^{\text{mono}}-D_{t}^{\text{pp}}}{D_{t}^{\text{pp}}},\frac{D_{t}^{\text{pp}}-D_{t}^{\text{mono}}}{D_{t}^{\text{mono}}})< \delta
\end{equation}

In addition, the monocular depth estimation is adjusted to align more closely with the planar-parallax module's depth prediction, as they should largely agree on static scenes. To prevent inaccuracies due to possible different scales, normalized depth $\overline{D}$ is used to ensure the alignment of the two outputs in defining $L_{\text{consist}}$ in~\cref{eqn:LOSS_consist}.

\begin{equation}
    \label{eqn:LOSS_consist}
     L_{consist} = \sum_{p_{i}}^{N} M_{static} | \overline{D_{t}^{mono}} - \overline{D_{t}^{pp}}|
\end{equation}

The training strategy is as follows: for the first 5 epochs, the total loss is $L_{mono}+L_{res}+L_{pp}$, when the pose is jointly optimized with these losses. Then, the homography loss $L_{homo}$ is added to correctly align the road. After 20 epochs, the planar-parallax module is frozen. Then, $L_{consist}$ is activated in addition to masking the $L_{mono}$ by $M_{static}$ to eliminate the dynamic objects. Hence, the full loss becomes $L_{mono}+L_{homo}+L_{consist}$. In addition, edge-aware smoothness loss is added during all the training for $D_{t}$ and $S_{t}$, as already utilized by most of the related work~\cite{manydepth,godard2019digging_monodepth2_selfsup,wang2023planedepth_selfsup,godard2017unsupervised_monodepth1}.

\section{Experiments}

We conducted a comprehensive evaluation of our MonoPP model is conducted, benchmarking it against the SOTA, specifically self-supervised monocular depth estimation methods. In addition, we included a baseline model that utilized the depth predicted from Monodepth2 to compute metric-scaled depth by the camera height information, scaling it directly with the height GT; more details are available in the supplementary materials. To further substantiate the efficacy of our approach, we conducted an ablation study. This allows us to demonstrate the significance of each component within our pipeline. For the purpose of comparison, we employ the standard metrics for depth estimation, as referenced in~\cite{eigen2014depth,eigen2015predicting}.

MonoPP is subjected to detailed testing on two datasets: KITTI~\cite{geiger2012kitti} and Cityscapes~\cite{cordts2016cityscapes}. Our findings reveal that our model delivers SOTA performance in terms of the predicted scaled depth in KITTI. Furthermore, MonoPP exhibits competitive results when assessed on a scale-invariant basis. Our analysis is more emphasized on KITTI, as to the best of our knowledge, there are no published self-supervised metric-scaled depth predictions on Cityscapes.
%Furthermore, MonoPP exhibits competitive results when assessed using scale-invariant metrics.

In \textbf{KITTI}~\cite{geiger2012kitti}, we employ the Eigen split~\cite{eigen2015predicting}. This split is traditionally utilized for single frame depth estimation. The KITTI Eigen test set includes 39,810 images for training and 4,424 for validation. We use the same intrinsics for all images and the camera position given in~\cite{geiger2012kitti}, which is parallel to the ground at a height of 1.65 meters.

In \textbf{Cityscapes}, our training process follows the methodologies outlined in~\cite{manydepth,yang2018lego,zhou2017unsupervised}. We utilize a total of 69,731 images from the monocular sequences. These images are preprocessed using the scripts derived from~\cite{zhou2017unsupervised}. Our evaluation is conducted on 1,525 test images, and we utilized the ground-truth derived from the disparity maps~\cite{hou2019multi} for evaluation. According to the calibration files of cityscapes, the camera mounting position is $\approx 1.2$ meters above the ground with a pitch of $2.18^{\circ}$~\cite{cordts2016cityscapes}.

\textbf{Implementation details}. We used the same augmentation, using the settings from~\cite{godard2017unsupervised_monodepth1,godard2019digging_monodepth2_selfsup}. The model is trained with an input and output resolution of 640x192, and we make use of $I_{t-1}$ and $I_{t+1}$ during training for the auto-masking strategy, similar to~\cite{godard2019digging_monodepth2_selfsup}. Also, we utilized Adam optimizer~\cite{kingma2014adam} for all the epochs with learning rate $10^{-4}$. For the hyper-parameters, $\alpha=0.85$, $\delta=0.2$, $\epsilon = 5$, and $\tau = \text{cos}(3^{o})$. To map the sigmoid output to the correct flowscales, we have used $f_{\text{min}}=-100$ and $f_{\text{max}}=100$ for the input resolution. The depth is capped at 80m to be comparable with other SOTA methods.

%\subsection{Results}
\textbf{Results}. Extensive quantitative analyses were conducted on state-of-the-art (SOTA) methods that utilize self-supervised learning. These methods do not require any ground-truth or guidance from synthetic data during the training process. A significant portion of recent research focuses on treating the depth task as a scale-invariant problem. Consequently, we divided our analysis into two categories: methods that scale predictions by the median scaling of the ground-truth~\cite{godard2019digging_monodepth2_selfsup,VADEPTH}, and methods that compare the predicted depth directly to the GT. The latter implies that the model’s output is a direct metric-depth estimation, as shown in~\cref{tab:kitti_tab}. Our model is competitive in metric-depth estimation, achieving results comparable to DynaDepth~\cite{dynadepth}, which uses pose information unlike MonoPP.

\begin{figure}[h]
  \centering
    \centering
    \includegraphics[width=0.98\linewidth]{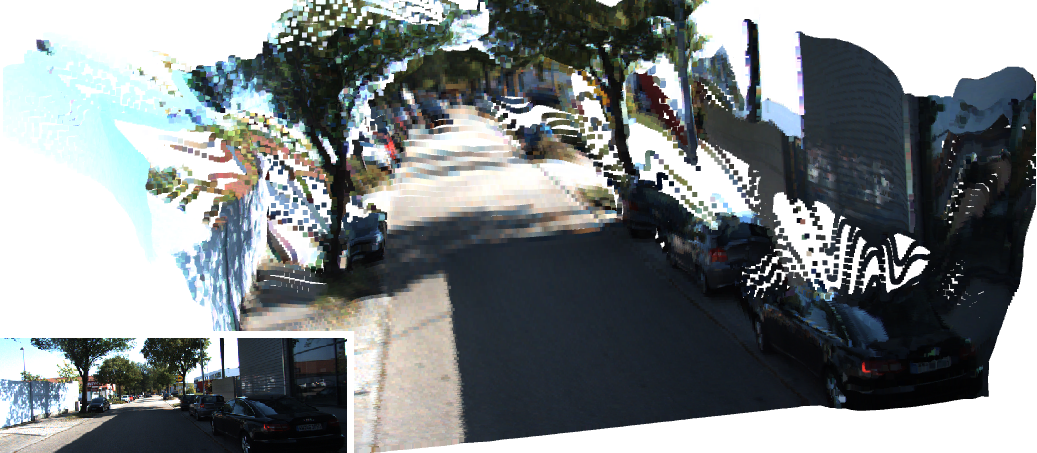}
  \caption{Rendered 3D point cloud from MonoPP on a KITTI Eigen split test sample (unseen during training). Input image is shown at the bottom left.}
  \label{fig:rendering_comb}
\end{figure}

In conclusion, MonoPP demonstrates superior performance using only the camera mounting position as a source for scale. It also achieves competitive results compared to methods that utilize pose information, which require an additional sensor. For example, a qualitative result in~\cref{fig:rendering_comb} illustrates the inference of a 3D point cloud from a single 2D image using the Eigen test split, rendered from a novel view angle. Furthermore, our inference model exhibits a latency of approximately 0.019 seconds per image when running on an Nvidia T4 GPU.
\begin{table*}[!ht]
\centering
\footnotesize % This line will decrease the font size
\setlength\tabcolsep{2 pt}
\renewcommand{\arraystretch}{1.15}
\begin{tabular*}{\textwidth}{@{\extracolsep{\fill}}|c|c|c|c|c|c|c|c|c|c|c|c|c|c|}
\hline

&$L_{\text{res}}$& $L_{\text{pp}}$& $L_{\text{consist}}$&$L_{\text{homo}}$&$M_{\text{static}}$&$M_{\text{cert}}$& Abs Rel $\downarrow$ & Sq Rel $\downarrow$ & RMSE $\downarrow$ & RMSE log $\downarrow$ & $\delta < 1.25$ $\uparrow$ & $\delta <1.25^2$ $\uparrow$ &  $\delta <1.25^3$ $\uparrow$ \\
\hline
MonoPP  (w/o $\theta_{pp}$) \label{row:1}& x&x&x&x&x&x&   0.238  &   1.387  &   6.047  &   0.326  &   0.346  &   0.912  &   0.966 \\
\hline
MonoPP& x&x&x&\checkmark&x&x&   0.124  &   0.950  &   5.009  &   0.203  &   0.853  &   0.952  &   0.979 \label{row:2}\\
\hline
MonoPP & x &x &\checkmark&\checkmark&x&x&   0.115  &   0.889  &   4.846  &   0.190  &   0.879  &   0.961  &   0.982   \label{row:3} \\
\hline
MonoPP & \checkmark &\checkmark &\checkmark&\checkmark&x&x&   0.112  &   0.947  &   4.932  &   0.188  &   0.886  &   0.961  &   0.982 \label{row:4}\\
\hline 
MonoPP & \checkmark &\checkmark &x&\checkmark&x&\checkmark&   0.110  &   0.903  &   4.840  &   0.187  &   0.889  &   0.962  &   0.982 \label{row:dagger_1}\\
\hline
MonoPP \textdagger & \checkmark &\checkmark &\checkmark&\checkmark&x&\checkmark&   0.115  &   1.030  &   5.148  &   0.191  &   0.880  &   0.960  &   0.981\label{row:6} \\
\hline
MonoPP & \checkmark &x &\checkmark&\checkmark&\checkmark&\checkmark&   0.109  &   0.915  &   4.844  &   0.188  &   0.890  &   0.961  &   0.981 \label{row:7}\\
\hline

MonoPP \textdagger & \checkmark &\checkmark &x&\checkmark&\checkmark&\checkmark&   0.109  &   0.910  &   4.830  &   0.187  &   0.890  &   0.962  &   0.981 \label{row:8_dagger_2} \\
\hline
MonoPP & \checkmark &\checkmark &x&\checkmark&\checkmark&\checkmark&   \textbf{0.107}  &   0.838  &   4.743  &   \textbf{0.185}  &   \textbf{0.891}  &   \textbf{0.962}  &   \textbf{0.982} \label{row:best} \\
\hline
MonoPP & \checkmark &\checkmark &\checkmark&\checkmark&\checkmark&\checkmark&   \textbf{0.107}  &   \textbf{0.835} &   \textbf{4.658} &   0.186  &   \textbf{0.891}  &  \textbf{0.962}  &   \textbf{0.982} \label{row:last}\\
\hline

%

% 0.231  &   1.958  &   6.653  &   0.255  &   0.710  &   0.944  &   0.979
\end{tabular*}

\caption{Ablation studies for different settings, performed on KITTI~\cite{geiger2012kitti} and evaluated on Eigen split~\cite{eigen2015predicting}, without GT median scaling. \textdagger~ this is the full model but without the further masking of $L_{\text{mono}}$ by $M_{\text{static}}$ in the final 5 epochs. $L_{\text{consist}}$ without $M_{\text{static}}$ assumes a true mask.}
\label{tab:ablations}
\end{table*}

We conducted a brief comparative analysis for Cityscapes~\cite{cordts2016cityscapes} in~\cref{tab:cityscape_quantative}. Notably, there are no published results on Cityscapes evaluation without the median scaling of GT. However, Kinoshita and Nishino~\cite{CameraHeightDoesnChange}, referred to as ``FUMET'' in~\cref{tab:cityscape_quantative}, reported re-training some models on Cityscapes and analyzing their performance without GT median scaling, as shown in~\cref{tab:cityscape_quantative}.

%\vspace{-0.4 cm}

\begin{table}[!ht]
  \centering
  {\footnotesize{
  \begin{tabularx}{\columnwidth}{|X|c|c|c|c|}
    \hline
    Method &  Abs Rel  &  Sq Rel  & RMSE  & $\delta < 1.25$   \\
    \hline
    PackNet~\textdagger~\cite{packnet_selfsup} & 0.504  & 6.639 & 14.90 & 0.029 \\
    
    VADepth~\textdagger~\cite{VADEPTH} & 0.363  & 7.115 & 11.95 & 0.295 \\
    FUMET~\cite{CameraHeightDoesnChange} & \textbf{0.125}  & \textbf{1.288} & \underline{6.359} & \underline{0.858} \\
       
    MonoPP &   \underline{0.135}  &  \underline{1.432}  &   \textbf{6.249}  &   \textbf{0.862}   \\
    MonoPP (ZS)&   0.216   &   3.156  & 12.113 & 0.580 \\
    \hline
  \end{tabularx}
  }}
  \caption{Quantitative results for Cityscapes~\cite{cordts2016cityscapes},~\textdagger~ these results are reported by~\cite{CameraHeightDoesnChange}. (ZS) means zero-shot testing by a model trained on KITTI. All results are evaluated without GT median scaling. }
  \label{tab:cityscape_quantative}
\end{table}

\begin{figure}[ht]
  \centering
  \begin{minipage}{.5\linewidth}
    \centering
    \includegraphics[width=0.98\linewidth]{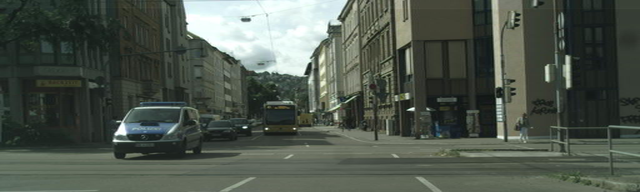}
    
    \includegraphics[width=0.98\linewidth]{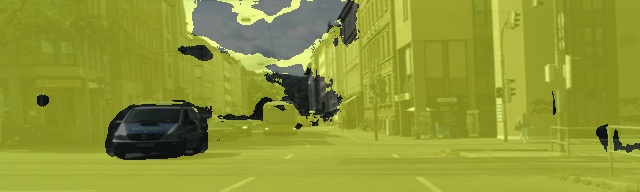}
    \subcaption{Example from Cityscapes~\cite{cordts2016cityscapes}}
  \end{minipage}%
  \begin{minipage}{.5\linewidth}
    \centering
    \includegraphics[width=0.98\linewidth]{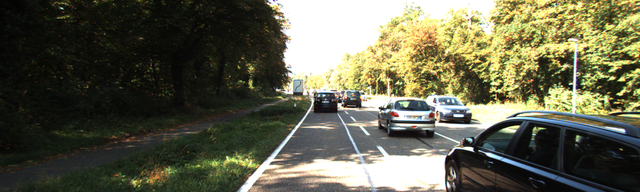}
    
    \includegraphics[width=0.98\linewidth]{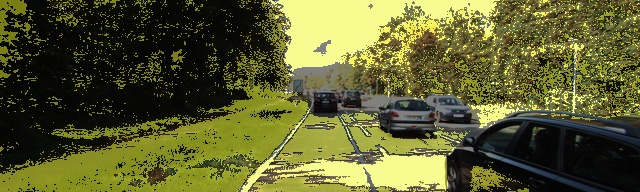}
    \subcaption{Example from KITTI~\cite{geiger2012kitti}}
  \end{minipage}
  \caption{Examples of masking the input images $I_{t}$, the first row is the input image , while the second is the masked image by $M_{\text{static}}$}
  \label{fig:staticmask_effect}
\end{figure}

\textbf{Ablation Studies}. In~\cref{tab:ablations}, we conducted comprehensive ablation studies to highlight the significance of each component, such as lossess and masks. As an example, training without $\theta_{\text{pp}}$ does not provide metric-scale information. However, utilizing $L_{\text{homo}}$ enforces metric-scale consistency and road planar alignment. Using $L_{\text{consist}}$ without $M_{\text{static}}$ and $M_{\text{cert}}$ leads to incorrect depth inheritance by $\theta_{\text{mono}}$ from $\theta_{\text{pp}}$. However, $L_{\text{consist}}$ does not have significant contribution when $M_{\text{static}}$ is used, as $M_{\text{static}}$ effectively distills knowledge as well. Nonetheless, we found $L_{\text{consist}}$ useful for fine-tuning on new datasets or specific cameras, aiding faster adaptation in fewer epochs. For larger datasets like KITTI or Cityscapes, $L_{\text{consist}}$ might be less crucial. In addition, $M_{\text{cert}}$ enhances the accuracy around the epipole area, which often contains dynamic objects or extreme depths. Additionally, using $M_{\text{static}}$ in the final 5 epochs to mask $L_{\text{mono}}$ prevents contamination by dynamic objects, as shown in~\cref{fig:staticmask_effect}.

\begin{figure}[h]
  \centering
  \begin{subfigure}{.5\columnwidth}
    \centering
    \includegraphics[width=0.98\linewidth]{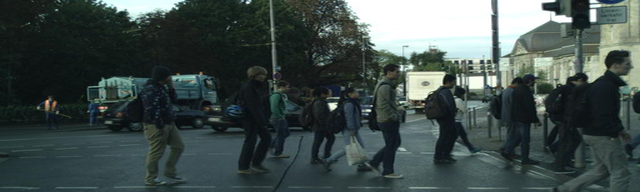}
    \caption{Input Image $I_{t}$}
    \label{fig:the_input_image_to_the_last_figure}
  \end{subfigure}%
  \begin{subfigure}{.5\columnwidth}
    \centering
    \includegraphics[width=0.98\linewidth]{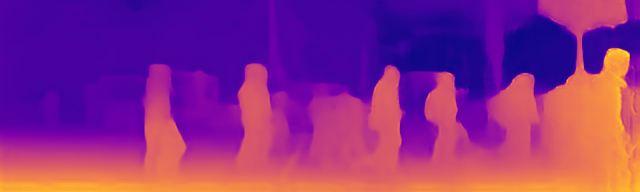}
    \caption{Predicted disparity by MonoPP}
    \label{fig:predicted_disparity_by_monoPP}
  \end{subfigure}
  
  \caption{An example from Cityscapes~\cite{cordts2016cityscapes}, where only a static frame is available (stationary vehicle) and a lot of dynamic objects.}
\label{fig:last_figure}
\end{figure}

Furthermore, training the same encoder with different decoder heads and objectives improves the encoder's image understanding and the student's performance, even without direct knowledge distillation from the teacher to the student. This is evidenced by the model trained with $L_{\text{pp}}$ and $L_{\text{res}}$ without $L_{\text{consist}}$ or $M_{\text{static}}$. This finding underscores that MonoPP is robust to some edge cases, as shown in~\cref{fig:last_figure}.

% \begin{table}[ht]
%   \centering
%   {\footnotesize{
%   \begin{tabularx}{\columnwidth}{|X|c|c|c|c|}
%     \hline
%     Method &  Abs Rel $\downarrow$ &  RMSE $\downarrow$ & RMSE log $\downarrow$ & $\delta < 1.25$ $\uparrow$  \\
%     \hline
%     w/o $L_{\text{res}}, L_{\text{pp}},L_{\text{consist\_pp}}$$ & Frumpy  & - & - & - \\
    
%     Yours & Frobbly  & - & - & - \\
%     Ours & x & - & -  & - \\
%     \hline
%   \end{tabularx}
%   }}
%   \caption{Ablation studies for different settings, performed on KITTI~\cite{geiger2012kitti} and evaluated on Eigen split~\cite{eigen2015predicting}}
%   \label{tab:example}
% \end{table}

% \vspace{-0.4 cm}

\section{Conclusion}

In conclusion, this work presents a metric-depth estimator that leverages video input and camera height, making it adaptable and practical. Unlike scale-invariant depth, which has limited use in real-world applications, our metric-depth estimator provides valuable, actionable data for a wide range of applications, without the need for costly ground-truth data. Our results highlight the potential of planar-parallax geometry in guiding self-supervised monocular depth estimation methods. This approach simplifies depth range definition based on flow, enhancing its adaptability to real-world scenarios. This represents a significant step forward in making depth estimation more intuitive and applicable across a myriad of real-world scenarios. In future work, we plan to evaluate the real-time performance of our model in an actual vehicle. This will involve utilizing a continuous signal for the camera’s mounting position and accounting for road slopes, providing more accurate and dynamic depth estimations.
%Additionally, we plan on using recent foundation models and fine-tuning them using our defined training strategy.
 
\clearpage

%%%%%%%%% REFERENCES
{\small
\newpage
\bibliographystyle{ieee_fullname}
\bibliography{wacv_bib}
}
\clearpage
\appendix
\section{Appendix}
\subsection{Computing the surface normal of the road}

In this section, we provide a detailed explanation of how the average normal vector, $$\mathbf{N}_{\text{avg}}(i, j)$$, is computed in our method. This process involves defining the central point and its neighbors, computing vectors to the neighbors, computing cross products of these vectors, and finally normalizing and averaging the results.

\begin{equation}
\mathbf{N}_{\text{avg}}(i, j) = \frac{1}{4} \sum_{k=1}^{4} \frac{\mathbf{V}_{k1}(i, j) \times \mathbf{V}_{k2}(i, j)}{\|\mathbf{V}_{k1}(i, j) \times \mathbf{V}_{k2}(i, j)\|}
\label{eqn:n_avg_append}
\end{equation}
where, ${V}_{k1}$ and ${V}_{k2}$ are the vectors from the central point to its neighbours, as shown in~\cref{eqn:vectors_k1k2} and~\cref{eqn:vectors_k1k2_2}.

\begin{equation}
\label{eqn:vectors_k1k2}
    \mathbf{V}_{k1}(i, j) = \mathbf{P}(i + \Delta i_k, j + \Delta j_k) - \mathbf{P}(i, j) 
\end{equation}
\begin{equation}
\label{eqn:vectors_k1k2_2}
    \mathbf{V}_{k2}(i, j) = \mathbf{P}(i + \Delta i_k', j + \Delta j_k') - \mathbf{P}(i, j)
\end{equation}

with \((\Delta i_k, \Delta j_k)\) and \((\Delta i_k', \Delta j_k')\) representing the offsets for the neighboring points. Then, the final surface normal for each pixel is the mean of these normals as shown in~\cref{eqn:n_avg_append}. For example, the computed vectors when using the nearest neighbors $n$ are:

\[
\begin{aligned}
\mathbf{V}_{11}(i, j) &= \mathbf{P}(i, j-n) - \mathbf{P}(i, j) \\
\mathbf{V}_{12}(i, j) &= \mathbf{P}(i-n, j) - \mathbf{P}(i, j) \\
\mathbf{V}_{21}(i, j) &= \mathbf{P}(i, j+n) - \mathbf{P}(i, j) \\
\mathbf{V}_{22}(i, j) &= \mathbf{P}(i+n, j) - \mathbf{P}(i, j) \\
\mathbf{V}_{31}(i, j) &= \mathbf{P}(i-n, j-n) - \mathbf{P}(i, j) \\
\mathbf{V}_{32}(i, j) &= \mathbf{P}(i+n, j-n) - \mathbf{P}(i, j) \\
\mathbf{V}_{41}(i, j) &= \mathbf{P}(i-n, j+n) - \mathbf{P}(i, j) \\
\mathbf{V}_{42}(i, j) &= \mathbf{P}(i+n, j+n) - \mathbf{P}(i, j)
\end{aligned}
\]
\textbf{Normalize and Average} We normalize each of the computed normal vectors and then take their average. This is done using~\cref{eqn:n_avg_append}. Then, this is the surface normal at this pixel. These normal vectors are filtered based on their alignment with the surface normal of the road, as described in our method.

\subsection{Auto-masking for re-projection loss}

In the context of self-supervised monocular depth estimation, auto-masking plays a crucial role in handling occlusions. The auto-masking mechanism is integrated into the minimum reprojection loss, which is defined as:

\begin{equation}
L_{\text{reproj}} = \min_{s} \text{pe}( \hat{I}_s , I_t)
\end{equation}

where \(I_s\) and \(I_t\) represent the source and target images respectively. The minimum operation in the loss function ensures that for each pixel in the target image, the model considers the best possible projection from the source images. This mechanism effectively serves as an automatic mask, enabling the model to be robust against occlusions. The pixels corresponding to occluded regions in the source image would have a high reprojection error, and hence, are automatically down-weighted in the loss computation. This auto-masking mechanism retains only the loss of pixels where the reprojection error of the warped image \(\hat{I}_{s}\) is lower than that of the original, unwarped source image \(I_{s}\). This can be mathematically represented as:

\begin{equation}
M_{\text{auto}} = [ \min_{s} \text{pe}(I_t, \hat{I}_{s}) < \min_{s} \text{pe}(I_t, I_{s})]
\end{equation}

where \(\text{pe}\) denotes the photometric error, \(I_t\) is the target image, \(\hat{I}_{s}\) is the image warped from \(s\) to \(t\), and \(I_{s}\) is the original, unwarped source image. The function \(M_{\text{auto}}\) serves as a mask that includes only the pixels where the reprojection error of the warped image is lower than that of the original image, $[ \hspace{3 pt}]$ denotes Iverson bracket.

where \(L_{\text{identity}}\) is the identity reprojection loss, \(L_{\text{reproj}}\) is the reprojection loss. The mask \(M\) takes the value 1 for pixels where the identity reprojection loss is less than the reprojection loss, and 0 otherwise. This effectively down-weights the contribution of occluded pixels in the loss computation, thereby making the model robust to occlusions. Also this mask $M_{\text{auto}}$ was utilized in the same fashion for masking out any invalid depth calculated by the teacher $\theta_{\text{pp}}$. 
\subsection{Smoothness loss}

\Cref{eqn:appendix_smoothness loss} is widely used in depth estimation models, which are often trainable methods. This equation encourages the disparity map to be smooth in regions where the image content is smooth, thereby reducing noise and improving the overall quality.

\begin{equation}
L_{smooth} = |\partial_x d^* t| \cdot e^{-|\partial_x I_t|} + |\partial_y d^* t| \cdot e^{-|\partial_y I_t|}
\label{eqn:appendix_smoothness loss}
\end{equation}

Where $d^{*}$ is the mean-normalized disparity. The exponential term makes this a robust function, meaning it is less sensitive to large disparity changes in the presence of strong image gradients, which may be gradients due to brightness changes, or any other external factors. This is important because edges in an image often correspond to depth discontinuities in the scene, so it is desirable for the disparity map to have sharp changes at these locations. Therefore, the smoothness loss helps to preserve edge information while ensuring overall smoothness, leading to more accurate and visually pleasing disparity maps.

\subsection{Scale computation for base-line comparison}
\label{sec:scale_recov}
In this section, we present the methodology employed to recover the scale from the depth map using surface normal vectors, which can be used for any scale-invariant model. Two different methods are used to compute the scale, both relying on the predicted height of the camera. Both methods rely on predicting the surface normal vectors from the depth map and then using those predictions to estimate the road plane and camera height.

\subsubsection{Method 1: Road Plane Estimation Using RANSAC}

The first method involves estimating the road plane by leveraging the surface normal vectors predicted from the depth map.

\begin{enumerate}
    \item \textbf{Depth Map and Surface Normals:} We start with the depth map of the scene, and from this, we predict the surface normal vectors for all the pixels in the image.
    \item \textbf{Identifying Flat Areas:} The flat areas in the scene (same as $M_{\text{flat}}$) are identified based on the surface normals, which are pointing almost in the same direction as the road surface normal.
    \item \textbf{Road Plane Estimation:} Using the predicted surface normals for the flat areas, we employ RANSAC to compute an estimate of the road plane. Although this robust estimation technique helps to exclude outliers, in some cases where there are a lot of non-road flat areas, it produces wrong results. 
    \item \textbf{Camera Height Estimation:} Once the road plane is estimated, the inferred height of the camera can be computed from its distance to the road plane.
    \item \textbf{Scale Adjustment:} The scale is then computed as the ratio between the predicted camera height and the actual known camera height.
\end{enumerate}

\subsubsection{Method 2: Median Height Estimation from All Pixels}

The second method is computationally more straightforward, as it avoids the RANSAC optimization process. Instead, it calculates the camera height by taking the median of the estimated heights from the flat pixels, providing a direct and efficient solution.

\begin{enumerate}
    \item \textbf{Depth Map and Surface Normals:} Similar to the first method, we start with the depth map and predict the surface normal vectors for all the pixels.
    \item \textbf{Height Calculation for the flat area:} We compute the inferred height of the camera using all flat-area pixels in the image.
    \item \textbf{Median Height Estimation:} Once the heights are computed for these pixels, we take the median of these inferred heights. The median serves as a robust estimate of the camera height, mitigating the influence of outliers.
    \item \textbf{Scale Adjustment:} Similar to the first method, we compute the scale as the ratio between the predicted median camera height and the actual known camera height.
\end{enumerate}

Both methods offer reliable approaches for recovering the scale based on the camera height derived from surface normals and depth data. Although these methods were not extensively tested, our results indicate that the second method is both simpler and yields more accurate scale estimates. For this reason, we adopted it in establishing our baseline using depth maps generated by Monodepth2.

\subsection{Zero-shot testing}

In the zero-shot testing scenario, we evaluated the model trained on KITTI using Cityscapes data, which the model had not seen during training. A key challenge in this process was the difference in scale between the datasets. To ensure a fair comparison, we implemented a straightforward module to adjust the scale by estimating the camera's height relative to the road. This adjustment was done using the same method described in the scale recovery process, ensuring consistent and accurate depth estimation across both datasets. 

\subsection{Qualitative results}
\begin{figure}[H]
  \centering
  
  \begin{subfigure}{.5\columnwidth}
    \centering
    \includegraphics[width=0.98\linewidth]{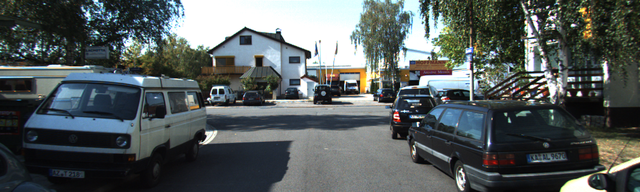}
    \caption{Input Image $I_{t}$}
  \end{subfigure}%
    \begin{subfigure}{.5\columnwidth}
    \centering
    \includegraphics[width=0.98\linewidth]{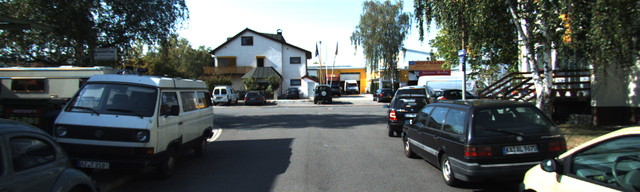}
    \caption{The source image, $I_{s}=I_{t-1}$}
  \end{subfigure}
    \begin{subfigure}{.5\columnwidth}
    \centering
    \includegraphics[width=0.98\linewidth]{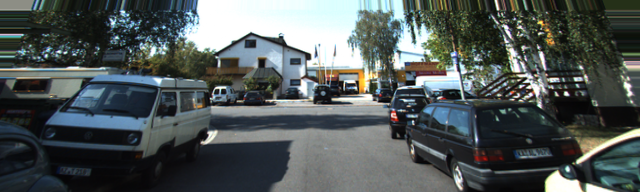}
    \caption{Input Image $I^{w}_{t-1}$}
  \end{subfigure}%
    \begin{subfigure}{.5\columnwidth}
    \centering
    \includegraphics[width=0.98\linewidth]{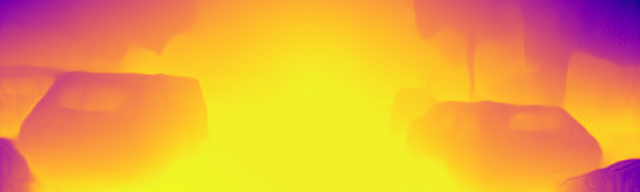}
    \caption{Flowscale output $s_{t}$}
  \end{subfigure}
  \begin{subfigure}{.5\columnwidth}
    \centering
    \includegraphics[width=0.98\linewidth]{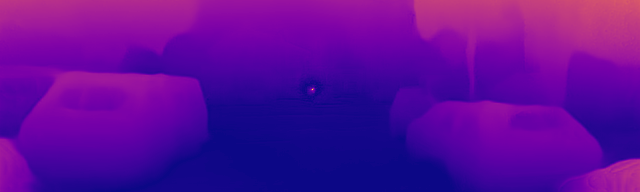}
    \caption{$\gamma = \frac{h}{d}$ }
  \end{subfigure}%
    \begin{subfigure}{.5\columnwidth}
    \centering
    \includegraphics[width=0.98\linewidth]{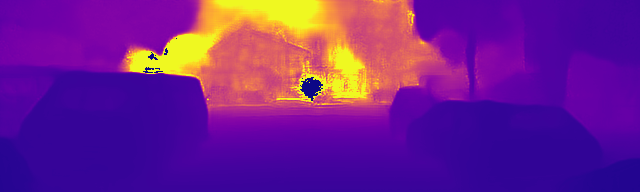}
    \caption{$D^{pp} * M_{\text{cert}}$, computed from $\theta_{pp}$}
  \end{subfigure}
    \begin{subfigure}{.5\columnwidth}
    \centering
    \includegraphics[width=0.98\linewidth]{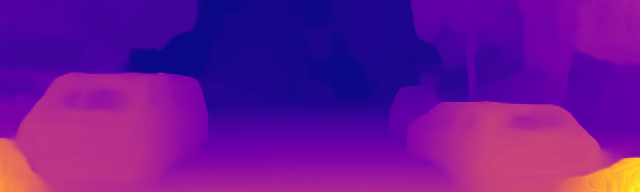}
    \caption{disparity, computed from $\theta_{mono}$}
  \end{subfigure}%
    \begin{subfigure}{.5\columnwidth}
    \centering
    \includegraphics[width=0.98\linewidth]{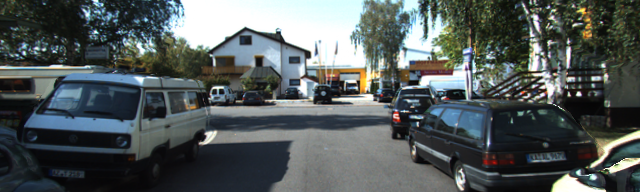}
    \caption{$I_{t}^{mono}$ computed from $D^{mono}$}
  \end{subfigure}
    \begin{subfigure}{.5\columnwidth}
    \centering
    \includegraphics[width=0.98\linewidth]{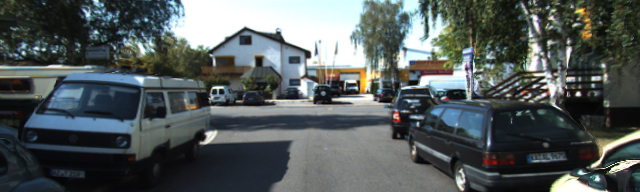}
    \caption{$I_{t}^{res}$ computed from $u_{res}$}
  \end{subfigure}%
      \begin{subfigure}{.5\columnwidth}
    \centering
    \includegraphics[width=0.98\linewidth]{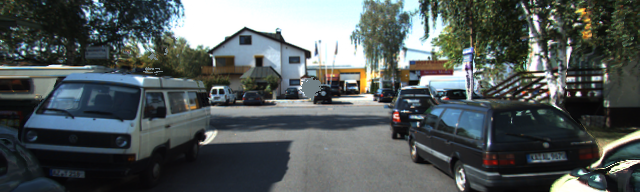}
    \caption{$I_{t}^{pp}$ computed from $D^{pp}$}
  \end{subfigure}

  \caption{Example for all the outputs as well as the intermediate outputs needed for computing the losses}
  \label{fig:full_pip}
\end{figure}

\begin{figure}[H]
  \centering
  \begin{subfigure}{.5\columnwidth}
    \centering
    \includegraphics[width=0.98\linewidth]{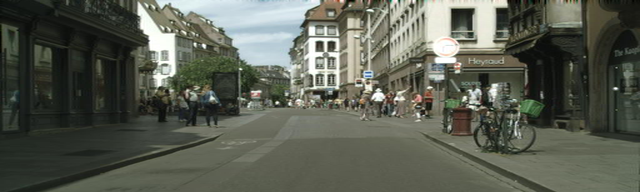}
    \caption{Input Image $I_{t}$}
  \end{subfigure}%
    \begin{subfigure}{.5\columnwidth}
    \centering
    \includegraphics[width=0.98\linewidth]{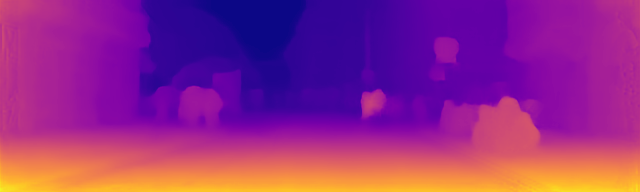}
    \caption{Disparity output}
  \end{subfigure}
    \begin{subfigure}{.5\columnwidth}
    \centering
    \includegraphics[width=0.98\linewidth]{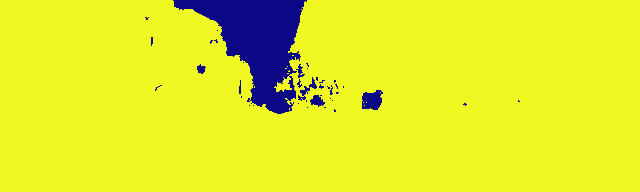}
    \caption{The static mask $M_{\text{static}}$}
  \end{subfigure}%
  \begin{subfigure}{.5\columnwidth}
    \centering
    \includegraphics[width=0.98\linewidth]{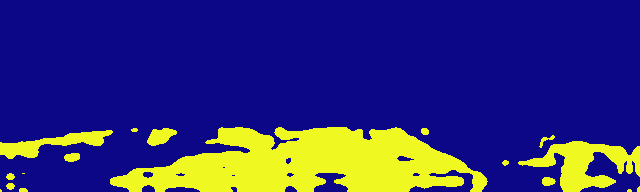}
    \caption{The flat area mask $M_{\text{flat}}$}
  \end{subfigure}

  \caption{A qualitative example from Cityscapes, which shows that $M_{\text{static}}$ will not affect the fully-static scene, the only masked area is the textureless sky, which is often mistaken for dynamic objects}
  \label{fig:good_cases_appendix_1}
\end{figure}

\begin{figure}[H]
  \centering
  \begin{subfigure}{.5\columnwidth}
    \centering
    \includegraphics[width=0.98\linewidth]{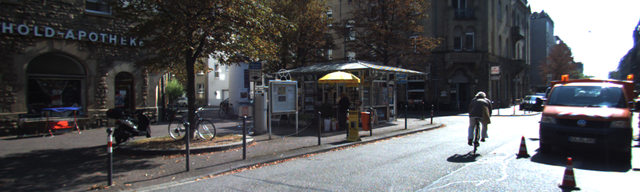}
    \caption{Input Image $I_{t}$}
  \end{subfigure}%
    \begin{subfigure}{.5\columnwidth}
    \centering
    \includegraphics[width=0.98\linewidth]{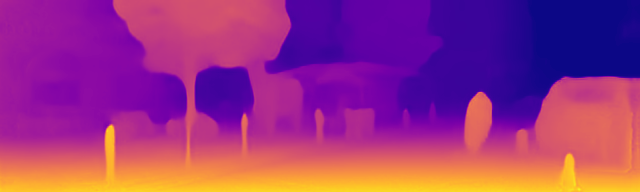}
    \caption{Disparity output}
  \end{subfigure}
    \begin{subfigure}{.5\columnwidth}
    \centering
    \includegraphics[width=0.98\linewidth]{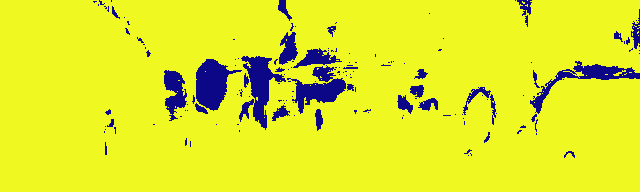}
    \caption{The static mask $M_{\text{static}}$}
  \end{subfigure}%
  \begin{subfigure}{.5\columnwidth}
    \centering
    \includegraphics[width=0.98\linewidth]{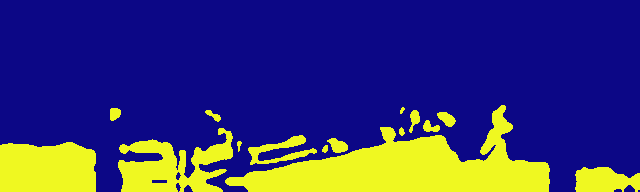}
    \caption{The flat area mask $M_{\text{flat}}$}
  \end{subfigure}
  
  \caption{Qualitative results on KITTI~\cite{geiger2012kitti}, the final depth result is correct. However,  it is a failure case, where $M_{\text{static}}$ classifies some static objects as dynamic and vice versa. This is happening sometimes due to the rotational movement of the vehicle, hence some objects are wrongly classified as dynamic objects.}
  \label{fig:good_cases_appendix_2}
  \end{figure}

\begin{figure}[H]
  \centering
  \begin{subfigure}{.5\columnwidth}
    \centering
    \includegraphics[width=0.98\linewidth]{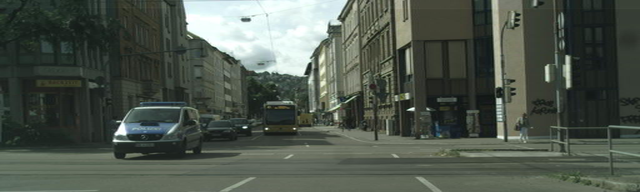}
    \caption{Input Image $I_{t}$}
  \end{subfigure}%
    \begin{subfigure}{.5\columnwidth}
    \centering
    \includegraphics[width=0.98\linewidth]{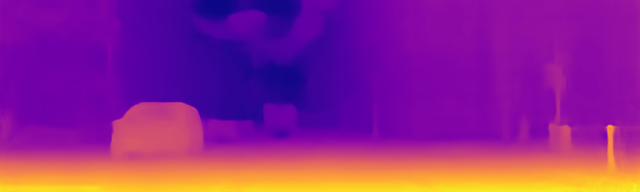}
    \caption{Disparity output}
  \end{subfigure}
    \begin{subfigure}{.5\columnwidth}
    \centering
    \includegraphics[width=0.98\linewidth]{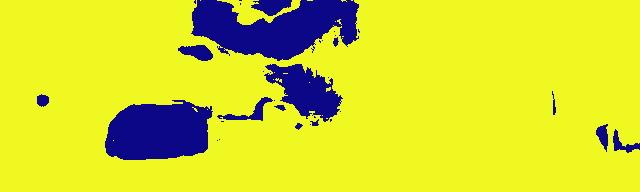}
    \caption{The static mask $M_{\text{static}}$}
  \end{subfigure}%
  \begin{subfigure}{.5\columnwidth}
    \centering
    \includegraphics[width=0.98\linewidth]{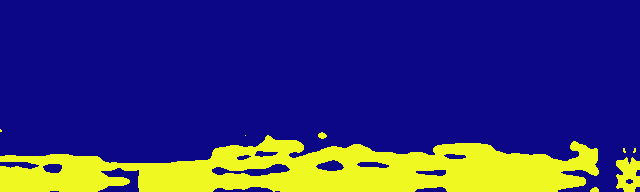}
    \caption{The flat area mask $M_{\text{flat}}$}
  \end{subfigure}
  
  \caption{Qualitative results on Cityscapes example, which was mentioned in the paper, and this is a good example of the usability of $M_{\text{static}}$.}
  \label{fig:good_cases_appendix_3}

\end{figure}

\begin{figure}[H]
  \centering
  \begin{subfigure}{.5\columnwidth}
    \centering
    \includegraphics[width=0.98\linewidth]{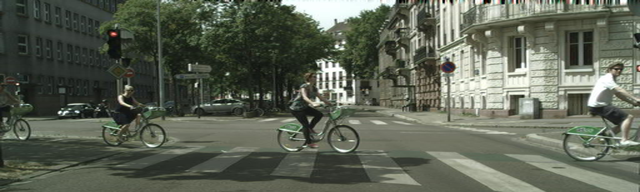}
    \caption{Input Image $I_{t}$}
  \end{subfigure}%
    \begin{subfigure}{.5\columnwidth}
    \centering
    \includegraphics[width=0.98\linewidth]{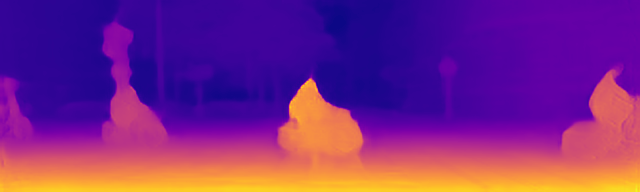}
    \caption{Disparity output}
  \end{subfigure}
    \begin{subfigure}{.5\columnwidth}
    \centering
    \includegraphics[width=0.98\linewidth]{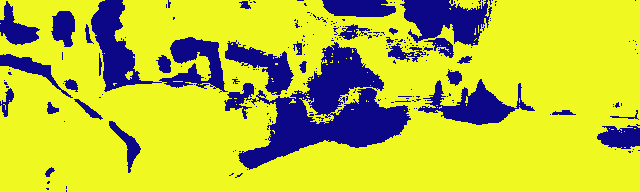}
    \caption{The static mask $M_{\text{static}}$}
  \end{subfigure}%
  \begin{subfigure}{.5\columnwidth}
    \centering
    \includegraphics[width=0.98\linewidth]{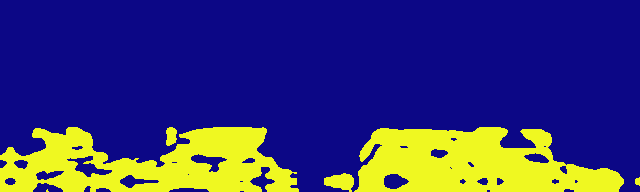}
    \caption{The flat area mask $M_{\text{flat}}$}
  \end{subfigure}
  \label{fig:last_figure_app}
  \caption{Qualitative results on Cityscapes, and this is one of the failure cases that $M_{\text{static}}$ filters out this dynamic object. However, it still was perceived as a bigger object, which is due to its closeness to the camera and its speed.}
    \label{fig:bad_cases_appendix_1}

\end{figure}

\begin{figure}[H]
  \centering
  \begin{subfigure}{.5\columnwidth}
    \centering
    \includegraphics[width=0.98\linewidth]{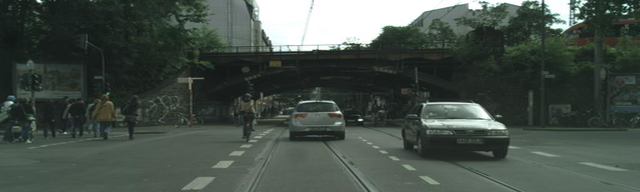}
    \caption{Input Image $I_{t}$}
  \end{subfigure}%
    \begin{subfigure}{.5\columnwidth}
    \centering
    \includegraphics[width=0.98\linewidth]{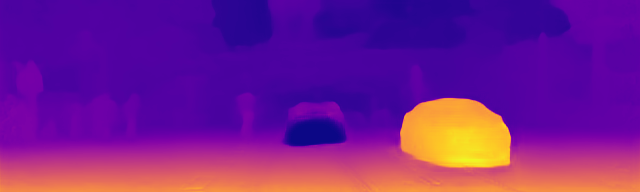}
    \caption{Disparity output}
  \end{subfigure}
    \begin{subfigure}{.5\columnwidth}
    \centering
    \includegraphics[width=0.98\linewidth]{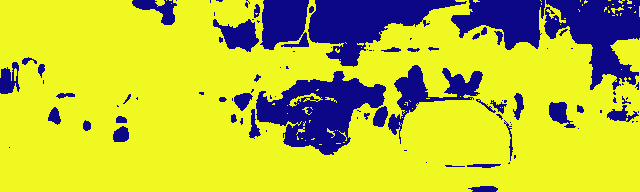}
    \caption{The static mask $M_{\text{static}}$}
  \end{subfigure}%
  \begin{subfigure}{.5\columnwidth}
    \centering
    \includegraphics[width=0.98\linewidth]{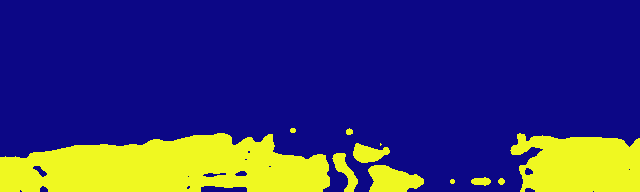}
    \caption{The flat area mask $M_{\text{flat}}$}
  \end{subfigure}
  \label{fig:last_figure_app_2}
  \caption{This is one of the challenging examples on cityscapes, which shows that of course our masking strategy does not filter out all dynamic objects. Hence, this will lead to hallucinated depth, which negatively affects our losses.}
      \label{fig:bad_cases_appendix_2}

\end{figure}

\begin{figure}[H]
  \centering
  \begin{subfigure}{.5\columnwidth}
    \centering
    \includegraphics[width=0.98\linewidth]{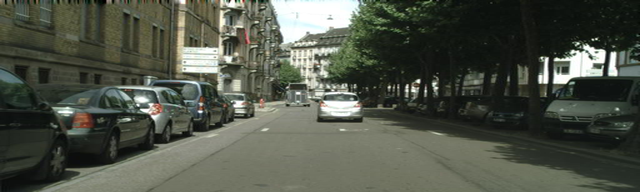}
    \caption{Input Image $I_{t}$}
  \end{subfigure}%
    \begin{subfigure}{.5\columnwidth}
    \centering
    \includegraphics[width=0.98\linewidth]{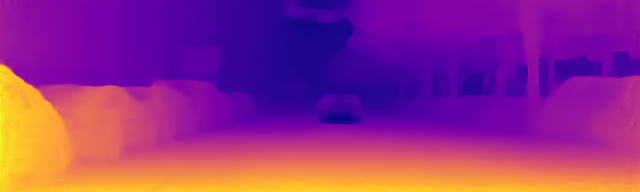}
    \caption{Disparity output}
  \end{subfigure}
    \begin{subfigure}{.5\columnwidth}
    \centering
    \includegraphics[width=0.98\linewidth]{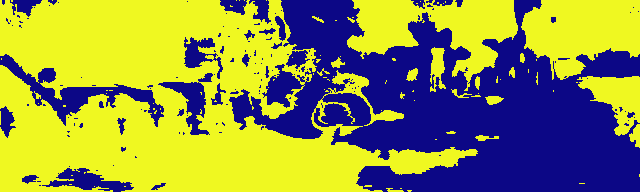}
    \caption{The static mask $M_{\text{static}}$}
  \end{subfigure}%
  \begin{subfigure}{.5\columnwidth}
    \centering
    \includegraphics[width=0.98\linewidth]{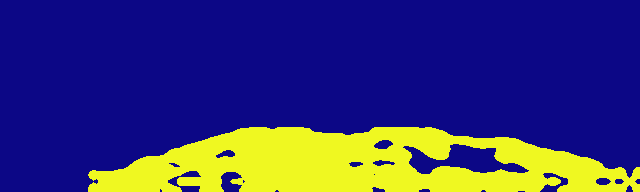}
    \caption{The flat area mask $M_{\text{flat}}$}
  \end{subfigure}
  \caption{This is one of the classic failures in Cityscapes, which is a moving car in front of the ego-vehicle, moving at similar speed and located around the epipole of the camera movement.}
\label{fig:bad_cases_appendix_3}

\end{figure}

In~\cref{fig:full_pip}, we illustrate the entire process of the model, encompassing both the teacher and student phases, during training. This serves as a practical demonstration of how each step is executed within the model. In~\cref{fig:good_cases_appendix_1,fig:good_cases_appendix_2,fig:good_cases_appendix_3}, we present successful cases where the output of $M_{\text{static}}$ effectively maps out dynamic objects. Conversely,~\cref{fig:bad_cases_appendix_1,fig:bad_cases_appendix_2,fig:bad_cases_appendix_3} highlight failure cases, where our masking strategy does not perform as intended, some of these failures are in masking the dynamic objects by $M_{\text{static}}$, as in~\cref{fig:bad_cases_appendix_2}, leading to completely incorrect depth as it calculates the depth based on the disparity of a moving objects. On the other hand, there are successful cases, such as in~\cref{fig:good_cases_appendix_2}, where the dynamic object is not entirely masked, but only its boundaries. Despite this, the disparity output for the dynamic object is still correct.

As shown in~\cref{fig:full_pip}, the training steps involved in our pipeline are outlined as follows. Starting with an input and source image, the teacher model computes the flow scale (epipolar flow scaling), followed by the calculation of the gamma parameter. The predicted depth is then masked by $M_{\text{cert}}$, ensuring that only reliable depth estimates are retained. Finally, novel views are synthesized using the outputs of the model.

\subsubsection{Good cases}
\begin{figure}[H]
  \centering
  \begin{subfigure}{.5\columnwidth}
    \centering
    \begin{tikzpicture}
      \node[anchor=south west,inner sep=0] (image) at (0,0) {\includegraphics[width=0.98\linewidth,height =0.3\linewidth]{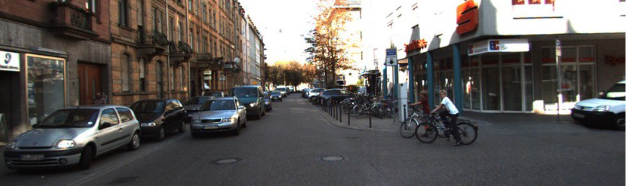}};
      \draw[red,ultra thick] (0.6\linewidth,0.21) rectangle (0.77\linewidth,0.2\linewidth); % Adjust these coordinates to position your box
            \draw[blue,ultra thick] (0.88\linewidth,0.21) rectangle (0.99\linewidth,0.2\linewidth);
    \end{tikzpicture}
    \caption{Original input}
    \label{fig:org_qual}
  \end{subfigure}%
    \begin{subfigure}{.5\columnwidth}
    \centering
    \begin{tikzpicture}
      \node[anchor=south west,inner sep=0] (image) at (0,0) {\includegraphics[width=0.98\linewidth,height =0.3\linewidth]{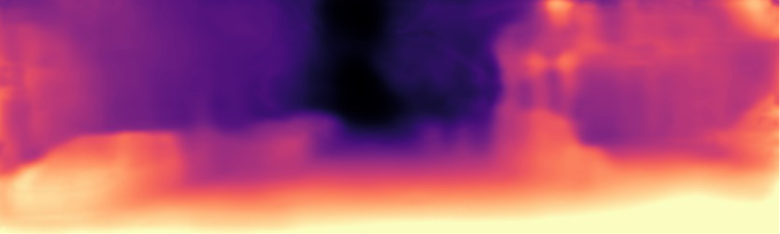}};
      \draw[red,ultra thick] (0.6\linewidth,0.21) rectangle (0.77\linewidth,0.2\linewidth); % Adjust these coordinates to position your box
            \draw[blue,ultra thick] (0.88\linewidth,0.21) rectangle (0.99\linewidth,0.2\linewidth);
    \end{tikzpicture}
    \caption{GeoNet (M)~\cite{yin2018geonet}}
    \label{fig:geonet_qual}
  \end{subfigure}
    \begin{subfigure}{.5\columnwidth}
    \centering
    \begin{tikzpicture}
      \node[anchor=south west,inner sep=0] (image) at (0,0) {\includegraphics[width=0.98\linewidth,height =0.3\linewidth]{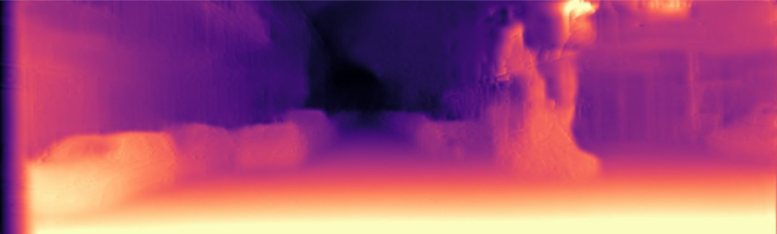}};
      \draw[red,ultra thick] (0.6\linewidth,0.21) rectangle (0.77\linewidth,0.2\linewidth); % Adjust these coordinates to position your box
            \draw[blue,ultra thick] (0.88\linewidth,0.21) rectangle (0.99\linewidth,0.2\linewidth);
    \end{tikzpicture}
    \caption{Monodepth1 (M)~\cite{godard2017unsupervised_monodepth1}}
    \label{fig:monodepth1_quali}
  \end{subfigure}%
    \begin{subfigure}{.5\columnwidth}
    \centering
    \begin{tikzpicture}
      \node[anchor=south west,inner sep=0] (image) at (0,0) {\includegraphics[width=0.98\linewidth,height =0.3\linewidth]{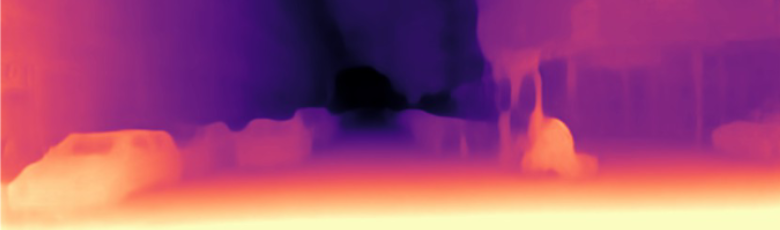}};
      \draw[red,ultra thick] (0.6\linewidth,0.21) rectangle (0.77\linewidth,0.2\linewidth); % Adjust these coordinates to position your box
            \draw[blue,ultra thick] (0.88\linewidth,0.21) rectangle (0.99\linewidth,0.2\linewidth);
    \end{tikzpicture}
    \caption{Monodepth2 (M)~\cite{godard2019digging_monodepth2_selfsup}}
    \label{fig:monodepth2_M_quali}
  \end{subfigure}
    \begin{subfigure}{.5\columnwidth}
    \centering
    \begin{tikzpicture}
      \node[anchor=south west,inner sep=0] (image) at (0,0) {\includegraphics[width=0.98\linewidth,height =0.3\linewidth]{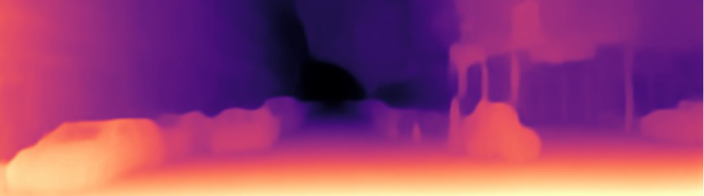}};
      \draw[red,ultra thick] (0.6\linewidth,0.21) rectangle (0.77\linewidth,0.2\linewidth); % Adjust these coordinates to position your box
            \draw[blue,ultra thick] (0.88\linewidth,0.21) rectangle (0.99\linewidth,0.2\linewidth);
    \end{tikzpicture}
    \caption{Monodepth2 (M+S)~\cite{godard2019digging_monodepth2_selfsup}}
    \label{fig:monodepth2_MS_quali}
  \end{subfigure}%
    \begin{subfigure}{.5\columnwidth}
    \centering
    \begin{tikzpicture}
      \node[anchor=south west,inner sep=0] (image) at (0,0) {\includegraphics[width=0.98\linewidth,height =0.3\linewidth]{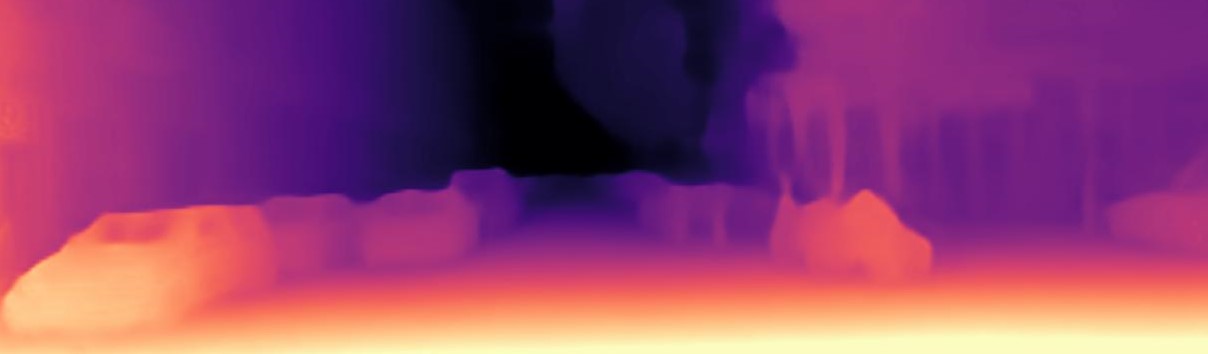}};
      \draw[red,ultra thick] (0.6\linewidth,0.21) rectangle (0.77\linewidth,0.2\linewidth);
      \draw[blue,ultra thick] (0.88\linewidth,0.21) rectangle (0.99\linewidth,0.2\linewidth);
    \end{tikzpicture}
    \caption{MonoPP (M)}
    \label{fig:sub1}
  \end{subfigure}
    
\caption{Qualitative results on KITTI~\cite{geiger2012kitti}, on eigen split~\cite{eigen2015predicting} in comparison with other SOTA methods. The finer details of the bike and the vehicle are detected.}  
\label{fig:compare_with_other_models}
\end{figure}
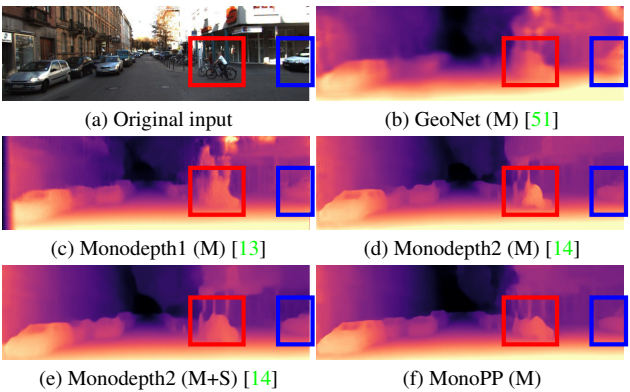

In~\cref{fig:compare_with_other_models}, we compare our model to recent approaches that also use single-frame monocular depth estimation. Although our model predicts metric-scaled depth, it achieves qualitatively comparable results to Monodepth2 (M+S), which was trained using stereo image pairs. This demonstrates that our approach performs competitively, despite relying solely on monocular input during training.

\subsubsection{Failure cases}

% \begin{figure*}[htbp]
%   \centering
%   \includegraphics[width=\textwidth]{your_image_file}
%   \caption{Your caption here}
%   \label{fig:your_label}
% \end{figure*}

\subsubsection{Rendered 3D point clouds results}

\begin{figure}[H]
  \centering
  \begin{subfigure}{0.98\columnwidth}
    \centering
    \includegraphics[width=0.98\linewidth]{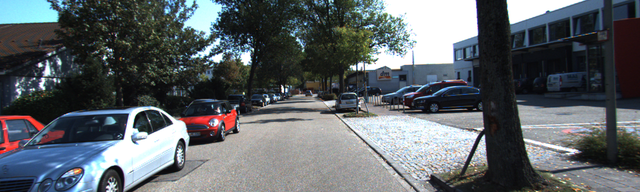}
  \end{subfigure}
    \begin{subfigure}{0.98\columnwidth}
    \centering
    \includegraphics[width=0.98\linewidth]{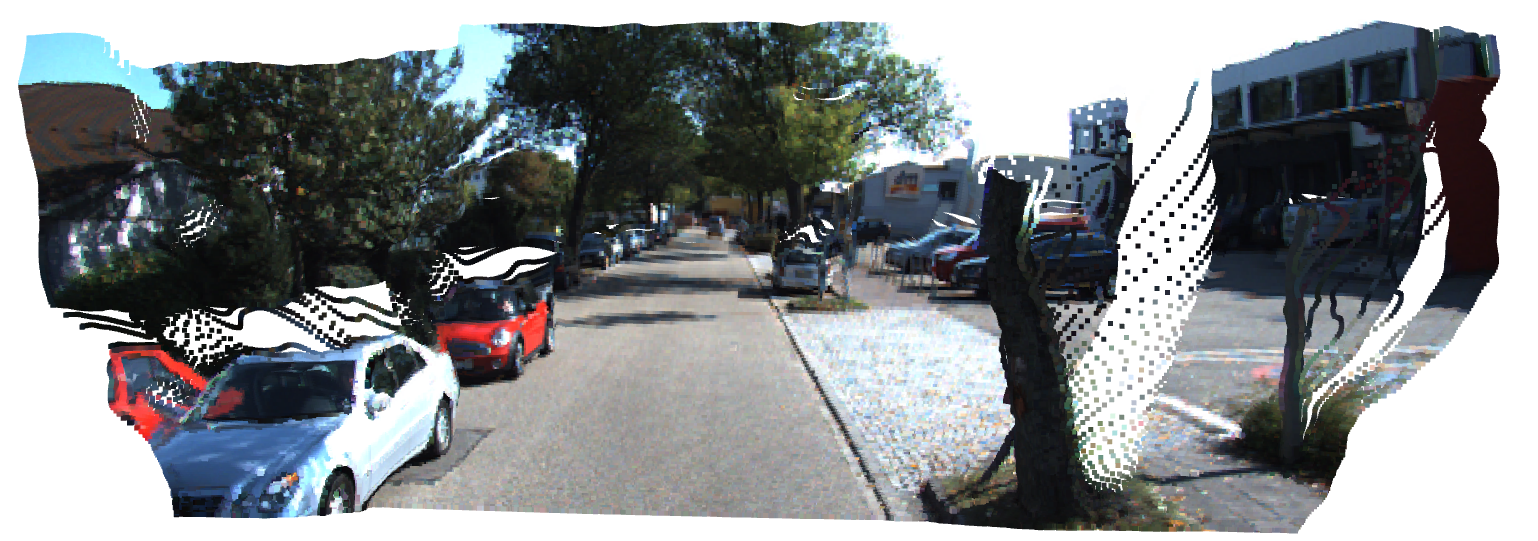}
  \end{subfigure}
    \begin{subfigure}{0.98\columnwidth}
    \centering
    \includegraphics[width=0.98\linewidth]{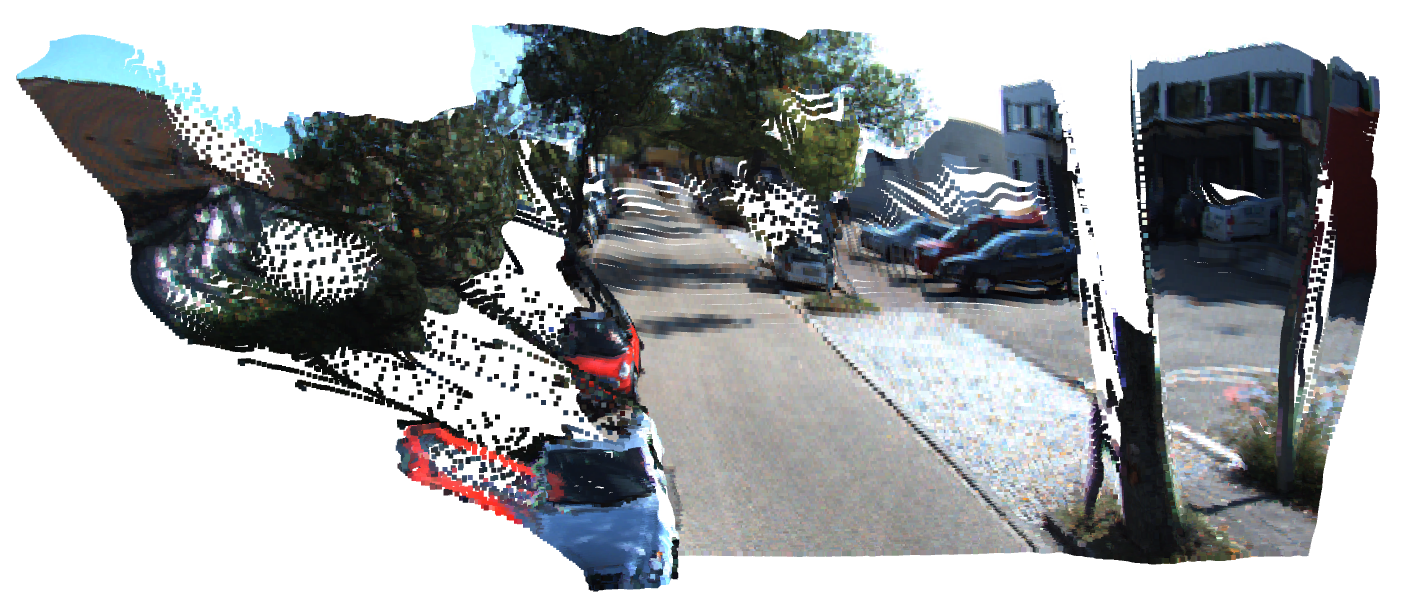}
  \end{subfigure}
  \begin{subfigure}{0.98\columnwidth}
    \centering
    \includegraphics[width=0.98\linewidth]{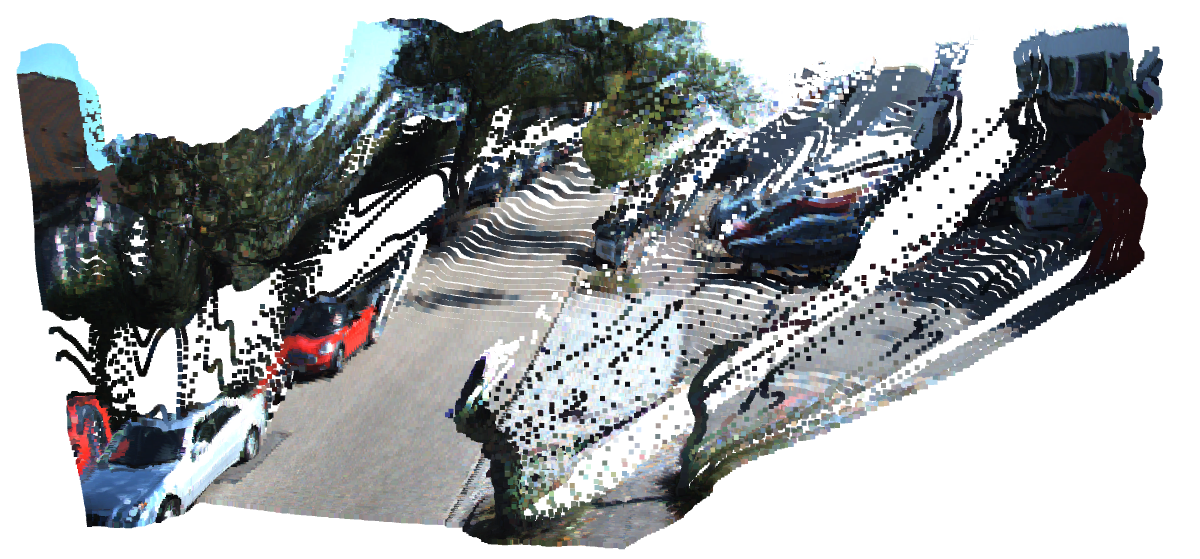}
  \end{subfigure}
  \caption{Rendered a 3D point cloud for KITTI data using MonoPP, based solely on a single 2D image input. Multiple view angles were used to visualize the scene.}
  \label{fig:rendered_app_1}
\end{figure}

\begin{figure}[h]
  \centering
  \begin{subfigure}{0.85\columnwidth}
    \centering
    \includegraphics[width=0.98\linewidth]{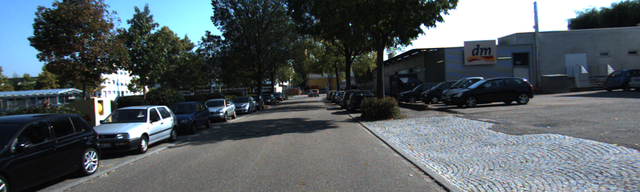}
  \end{subfigure}
    \begin{subfigure}{.85\columnwidth}
    \centering
    \includegraphics[width=0.98\linewidth]{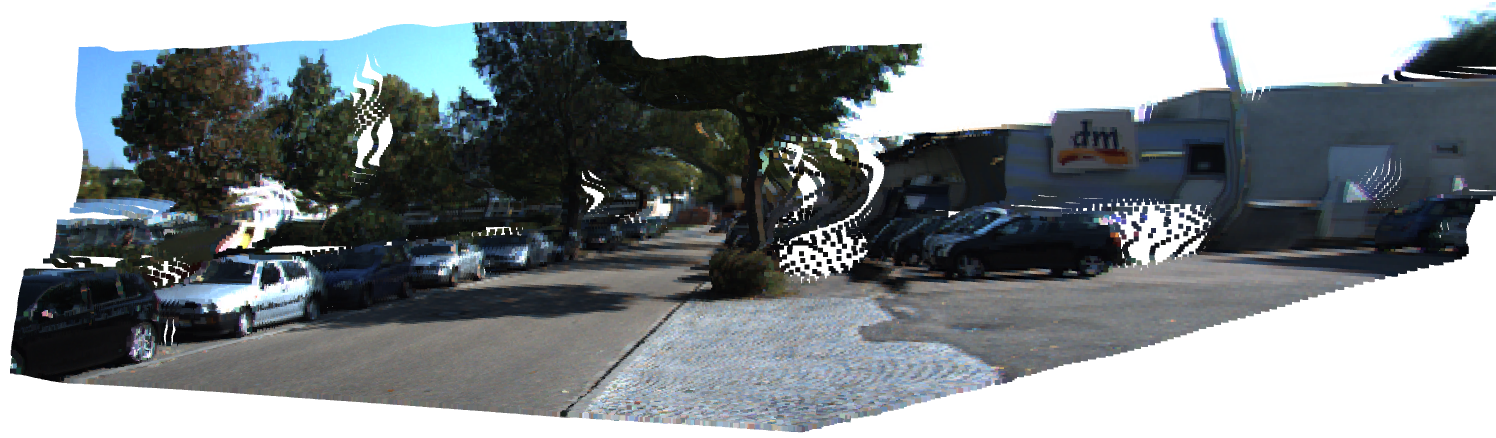}
  \end{subfigure}
    \begin{subfigure}{.85\columnwidth}
    \centering
    \includegraphics[width=0.98\linewidth]{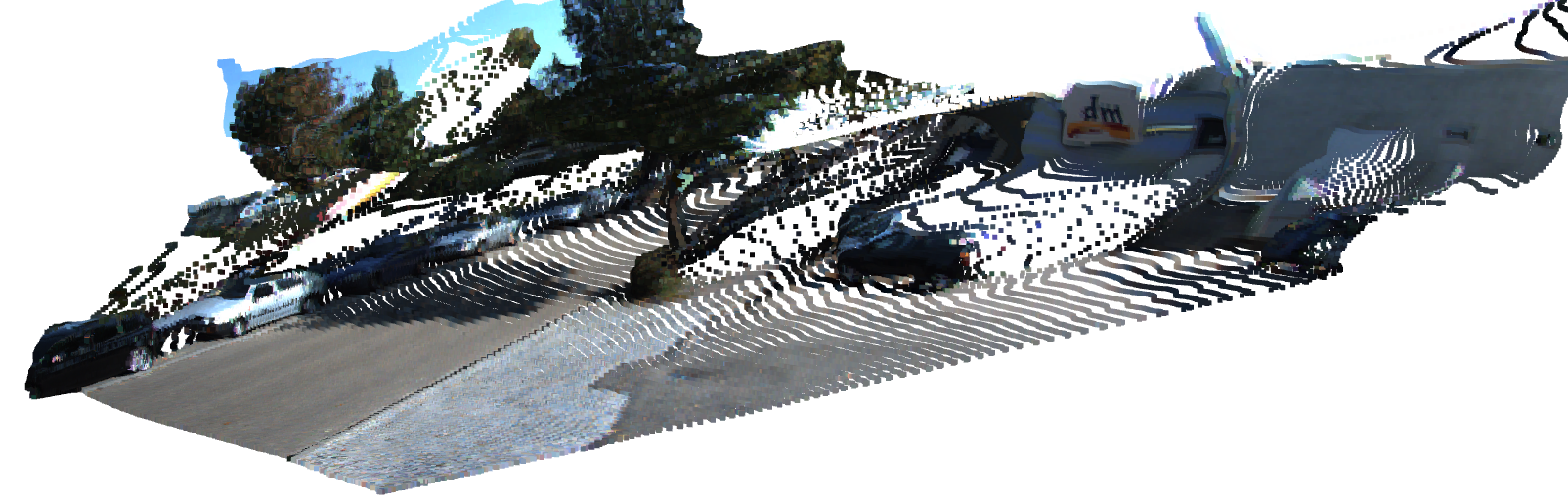}
  \end{subfigure}
      \begin{subfigure}{.85\columnwidth}
    \centering
    \includegraphics[width=0.98\linewidth]{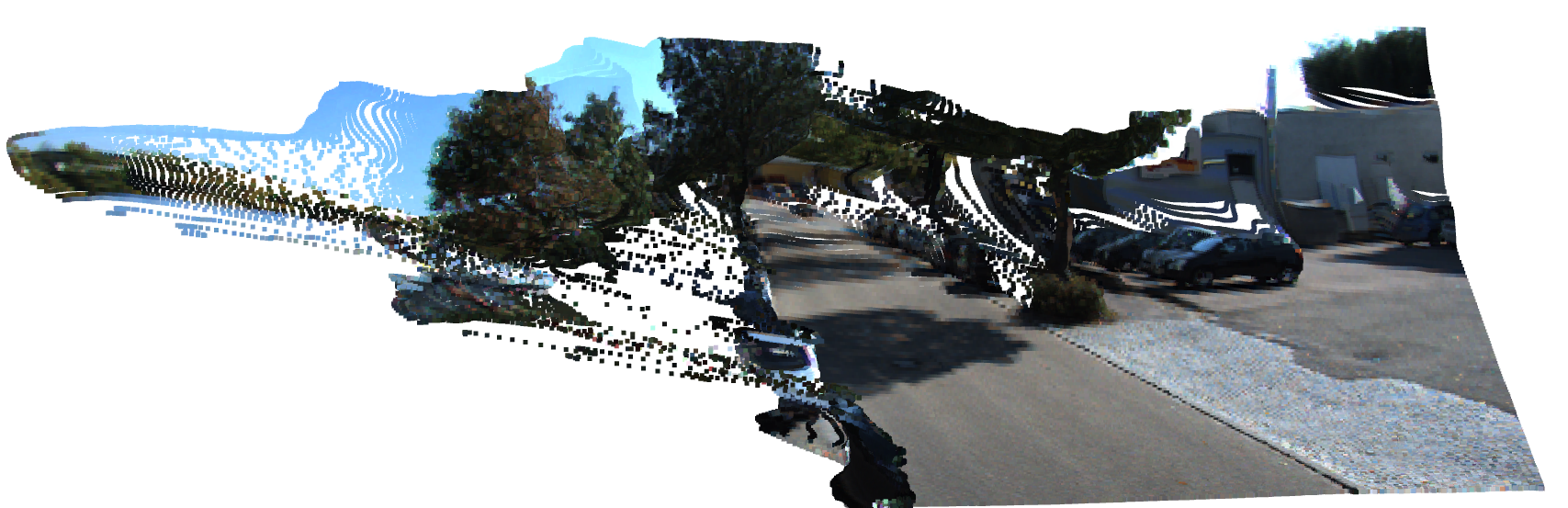}
  \end{subfigure}
  \caption{An additional example shows interesting faraway reconstruction of the rendered point clouds from a single image using MonoPP}
  \label{fig:rendered_app_2}

\end{figure}

\begin{figure}[h]
  \centering
  \begin{subfigure}{0.85\columnwidth}
    \centering
    \includegraphics[width=0.98\linewidth]{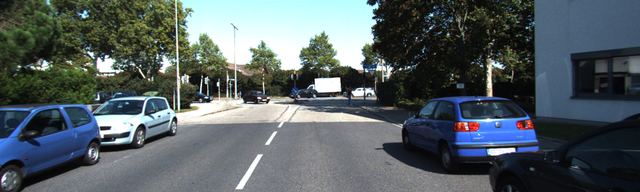}
  \end{subfigure}
    \begin{subfigure}{0.85\columnwidth}
    \centering
    \includegraphics[width=0.98\linewidth]{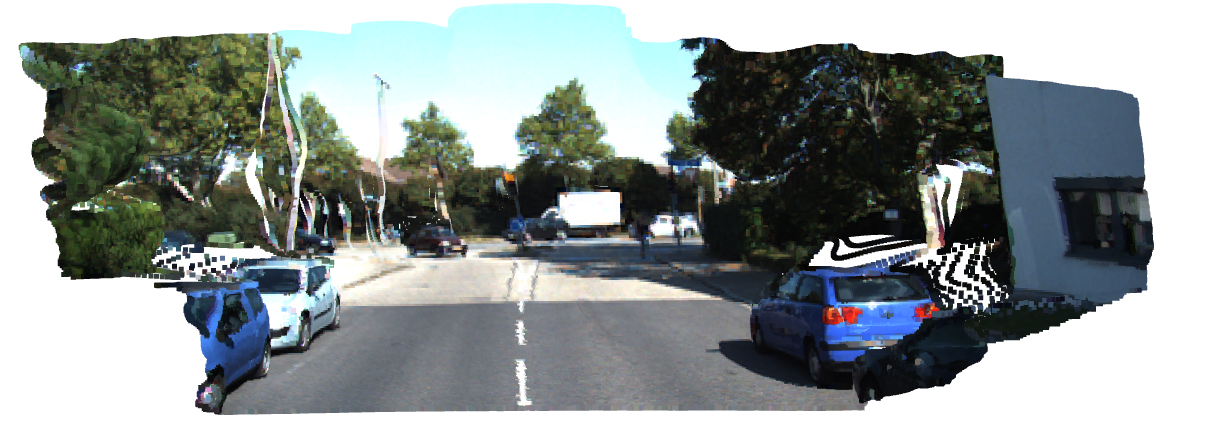}
  \end{subfigure}
    \begin{subfigure}{0.85\columnwidth}
    \centering
    \includegraphics[width=0.98\linewidth]{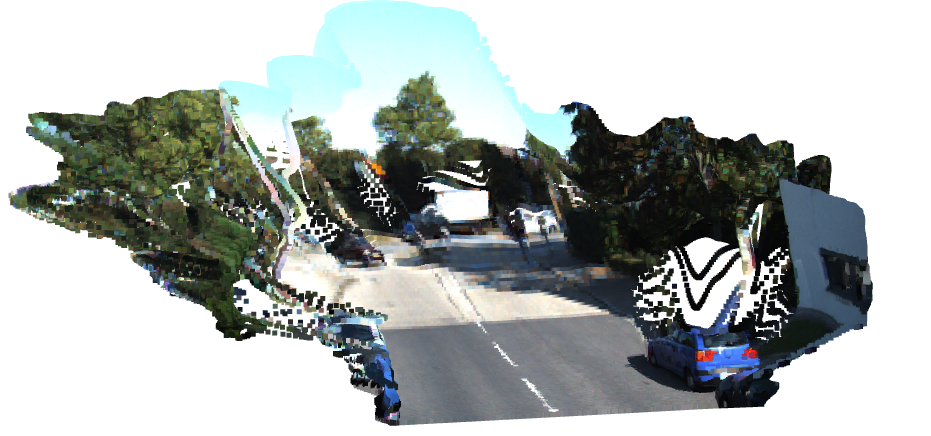}
  \end{subfigure}
  \caption{This example shows a good quality of rendered scene from a different view angle}
  \label{fig:rendered_app_3}

\end{figure}

\begin{figure}[h]
  \centering
  \begin{subfigure}{0.98\columnwidth}
    \centering
    \includegraphics[width=0.98\linewidth]{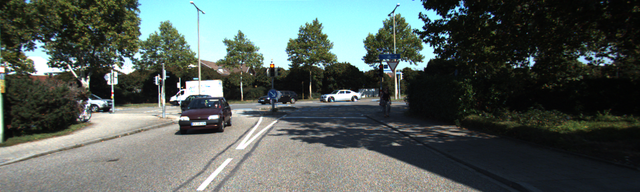}
  \end{subfigure}
    \begin{subfigure}{0.98\columnwidth}
    \centering
    \includegraphics[width=0.98\linewidth]{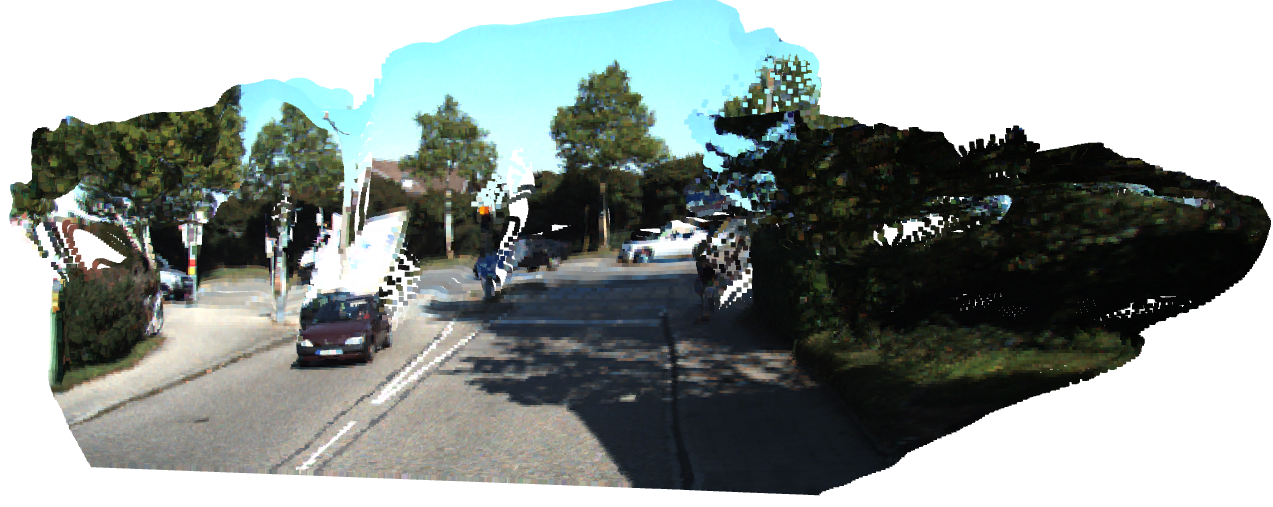}
  \end{subfigure}
    \begin{subfigure}{0.98\columnwidth}
    \centering
    \includegraphics[width=0.98\linewidth]{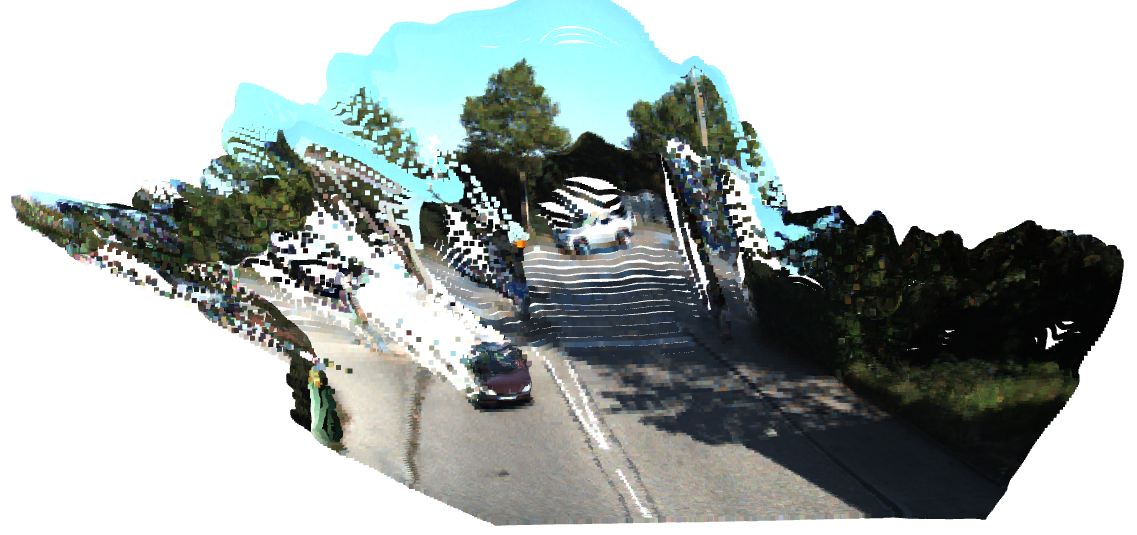}
  \end{subfigure}
  % \begin{subfigure}{0.98\columnwidth}
  %   \centering
  %   \includegraphics[width=0.98\linewidth]{images/rendering/7_dynamic_x/3.png}
  % \end{subfigure}
  %   \begin{subfigure}{0.98\columnwidth}
  %   \centering
  %   \includegraphics[width=0.98\linewidth]{images/rendering/7_dynamic_x/5.png}
  % \end{subfigure}
  \caption{This is a special sample which contains a lot of moving dynamic objects, which are more prone to error. However, the rendered scenes are of good quality}
  \label{fig:rendered_app_4}

\end{figure}

\begin{figure}[h]
  \centering
  \begin{subfigure}{0.98\columnwidth}
    \centering
    \includegraphics[width=0.98\linewidth]{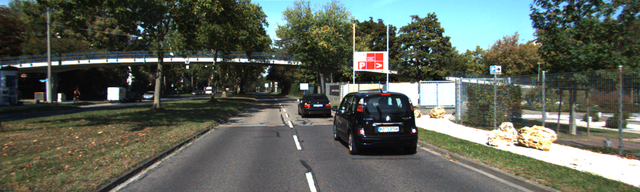}
  \end{subfigure}
    \begin{subfigure}{0.98\columnwidth}
    \centering
    \includegraphics[width=0.98\linewidth]{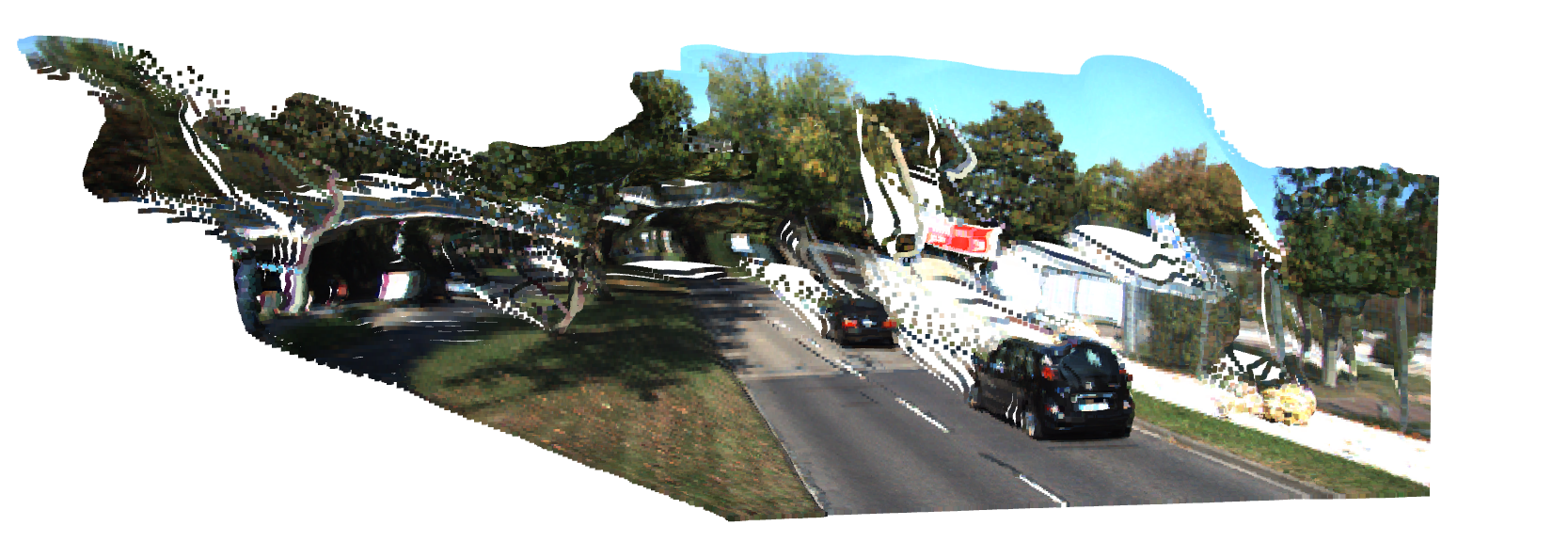}
  \end{subfigure}
    \begin{subfigure}{0.98\columnwidth}
    \centering
    \includegraphics[width=0.98\linewidth]{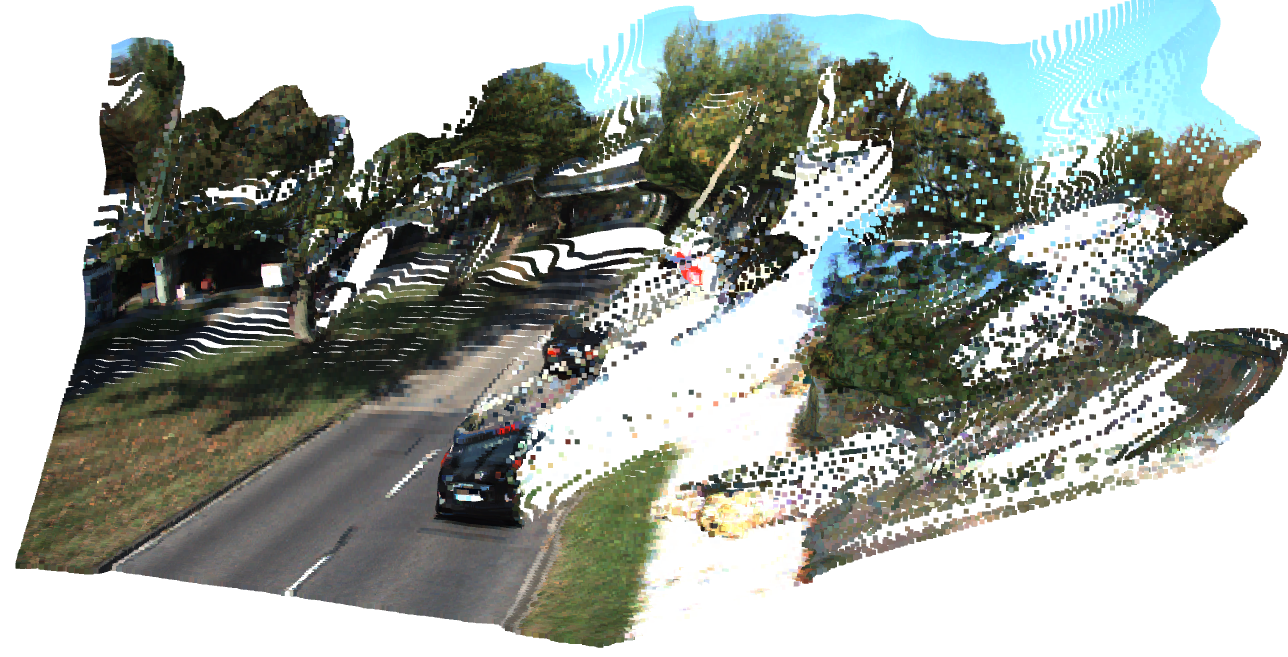}
  \end{subfigure}
  % \begin{subfigure}{0.98\columnwidth}
  %   \centering
  %   \includegraphics[width=0.98\linewidth]{images/rendering/7_dynamic_x/3.png}
  % \end{subfigure}
  %   \begin{subfigure}{0.98\columnwidth}
  %   \centering
  %   \includegraphics[width=0.98\linewidth]{images/rendering/7_dynamic_x/5.png}
  % \end{subfigure}
  \caption{This is a special sample which contains a lot of moving dynamic objects, which are more prone to error. However, the rendered scenes are of good quality}
  \label{fig:rendered_app_5}

\end{figure}

All the presented figures,~\cref{fig:rendered_app_1,fig:rendered_app_2,fig:rendered_app_3,fig:rendered_app_4,fig:rendered_app_5}, are formatted such as the first top image is the input to the inference network for depth prediction, and then the 3D point clouds are rendered from this single image only. All the examples are samples from the evaluation Eigen-split benchmark of KITTI dataset, which means that the network was not trained on these samples. 

\subsection{Quantitative results for KITTI}
~\cref{tab:kitti_tab_appendix} provides a comprehensive overview of the state-of-the-art (SOTA) methods in the field of self-supervised monocular depth estimation (with and without GT median scaling). It delineates the key differences between single-frame and multi-frame methods, providing valuable insights into their respective strengths and limitations. The table serves as a useful resource for future researchers, as it underscores the general superiority of multi-frame methods in terms of performance. However, it also highlights an important caveat: in scenarios where there is no baseline available, \ie only a single frame is available, single-frame methods may offer better results. This analysis can guide future research in this domain, informing the choice of methods based on the specific constraints.

\clearpage
\begin{table*}[ht]
\centering
\footnotesize % This line will decrease the font size
\setlength\tabcolsep{2.5 pt}
\renewcommand{\arraystretch}{1.15}
\begin{tabular*}{\textwidth}{@{\extracolsep{\fill}}|c|c|c|c|c|c|c|c|c|c|c|c|}
\hline
 & Year & Method & Test frames & Train  & Abs Rel $\downarrow$ & Sq Rel $\downarrow$ & RMSE $\downarrow$ & RMSE log $\downarrow$ & $\delta < 1.25$ $\uparrow$ & $\delta <1.25^2$ $\uparrow$ &  $\delta <1.25^3$ $\uparrow$ \\
\cline{2-12}    
  \multirow{17}{*}{\rotatebox[origin=c]{90}{scaled by GT}} & 2017 & Monodepth1~\cite{godard2017unsupervised_monodepth1} & 1 & M & 0.148 & 1.344  & 5.927 & 0.247 & 0.803 & 0.922 & 0.964 \\
 \cline{2-12}
 & 2018 & GeoNet~\cite{yin2018geonet} & N &  M & 0.149 & 1.060 & 5.567 & 0.226 & 0.796 & 0.935 & 0.975 \\
\cline{2-12}

 & \multirow{2}{*}{2019} & \multirow{2}{*}{Monodepth2~\cite{godard2019digging_monodepth2_selfsup}} & \multirow{2}{*}{1} & M & 0.115 & 0.903 & 4.863 & 0.193 & 0.877 & 0.959 & 0.981 \\
 \cline{5-12}

 &  &  & & M+S & 0.106 & 0.818 & 4.750 & 0.196 & 0.874 & 0.957 & 0.979 \\
 \cline{2-12}
 
 & 2020 & Patil~\etal~\cite{patil2020don} & N &  M & 0.11 & 0.82 & 4.65 & 0.187 & 0.883 & 0.961 & 0.982 \\
\cline{2-12}

 & 2020 & PackNet-SFM~\cite{packnet_selfsup} & 1 &  M & 0.111 & 0.785 & 4.601 & 0.189 & 0.878 & 0.960 & 0.982 \\
\cline{2-12}

 & 2020 & DNet~\cite{DNet} & 1 &  M & 0.113 & 0.864 & 4.812 & 0.191 & 0.877 & 0.960 & 0.981 \\
\cline{2-12}

 & \multirow{2}{*}{2021} & \multirow{2}{*}{ManyDepth~\cite{manydepth}} & N & M & 0.098  & 0.770  & 4.459  & 0.176   & 0.90   & \underline{0.965}  & \underline{0.983} \\
 \cline{4-12}

 &  &  & 1 & M & 0.106 & 0.818 & 4.750 & 0.196 & 0.874 & 0.957 & 0.979 \\
 \cline{2-12}
 & 2021 & CADepth~\cite{CADepth_2021} & 1 &  M & 0.110 & 0.812 & 4.686 & 0.187 & 0.882 & 0.961 & 0.981 \\
\cline{2-12}
 & 2021 & Sui~\etal ~\cite{roadaware_SFM_2021_selfsup_scaled} & 1 &  M & 0.111  &0.894 & 4.779 & 0.189 & 0.883  &0.960 &0.981 \\
\cline{2-12}

 & 2022 & VADepth~\cite{VADEPTH} & 1 &  M & 0.104  &0.774 &4.552 & 0.181 & 0.892  &0.965 & \underline{0.983} \\
\cline{2-12}
 & 2022 & MonoFormer~\cite{monoformer} & 1 &  M & 0.108  &0.806 &4.594 & 0.184 & 0.884  &0.963 & \underline{0.983} \\
\cline{2-12}
 & 2022 & DepthFormer~\cite{guizilini2022multiframe_toyota} & N &  M & \textbf{0.090}  & \textbf{0.661}  & \textbf{4.149} & \underline{0.175} & \underline{0.905}  & \textbf{0.967} & \textbf{0.984} \\
\cline{2-12}
 & 2022 & MonoViT~\cite{zhao2022monovit} & 1 &  M & 0.099 & 0.708  & 4.372 & \underline{0.175} & 0.900 &\textbf{0.967} & \textbf{0.984} \\
\cline{2-12}

 & 2023 & Lite-Mono~\cite{litemono_2023_selfsup} & 1 &  M & 0.107  &0.765 &4.561 & 0.183 & 0.886  &0.963 & \underline{0.983} \\
\cline{2-12}

 & 2023 & Lite-Mono-S~\cite{litemono_2023_selfsup} & 1 &  M & 0.110  &0.802 &4.671 & 0.186 & 0.879  &0.961 &0.982 \\
\cline{2-12}
& 2023 & TriDepth~\cite{tridepth}  & 1 &  M & \underline{0.093}  & \underline{0.665} & \underline{4.272} & \textbf{0.172} & \textbf{0.907}  & \textbf{0.967} &\textbf{0.984} \\
 \cline{2-12} 
 &  \multicolumn{2}{c|}{MonoPP (ours) } & 1 &  M & 0.105  &0.776 &4.640 & 0.185 & 0.891   &0.962 &0.982 \\
\hline
\hline
 \multirow{8}{*}{\rotatebox[origin=c]{90}{w/o scaling}}  & 2019 & Monodepth2**~\cite{godard2019digging_monodepth2_selfsup} & 1 &  camH &   0.126  &   0.973  &   4.880  &   0.198  &   0.864  &   0.957  &   0.980  \\
\cline{2-12}

& 2020 & DNet~\cite{DNet} & 1 &  M+camH & 0.118  &0.925 &  4.918  & 0.199 &0.862 &0.953  &0.979 \\
\cline{2-12}
 & 2020 & Zhao~\etal ~\cite{depth_by_GANS} & 1 &  M+SC &  0.146  &1.084 &5.445 & 0.221 & 0.807  &0.936 &0.976\\
\cline{2-12}
 & 2020 & PackNet~\cite{packnet_selfsup} & 1 &  M+V &  0.111  &0.829 &4.788 & 0.199 & 0.864  &0.954 &0.980\\
\cline{2-12}
 & 2021 & Wagstaff~\etal ~\cite{wagstaff_scalerecovery_2021_SELFSUP} & 1 &  M+Pose &  0.123  &0.996 &5.253 & 0.213 & 0.840  &0.947 &0.978 \\
\cline{2-12}
 & 2021 & Wagstaff~\etal ~\cite{wagstaff_scalerecovery_2021_SELFSUP} & 1 &  M+camH &  0.155  &1.657 &5.615 & 0.236 & 0.809  &0.924 &0.959 \\
\cline{2-12}

& 2021 & Sui~\etal~\cite{roadaware_SFM_2021_selfsup_scaled} & 1 &  M+camH & 0.128  &0.936 &  5.063  & 0.214 & 0.847 &0.951  &0.978 \\
\cline{2-12}
 & 2022 & VADepth~\cite{VADEPTH} & 1 &  M+camH &  0.109  & \underline{0.785} & \underline{4.624} & 0.190 & 0.875  & \underline{0.960} & \textbf{0.982} \\
 \cline{2-12}
  & 2022 & DynaDepth~\cite{dynadepth} & 1 &  M+Pose &  \underline{0.108}  & \textbf{0.761} & \textbf{4.608} & \underline{0.187} & \underline{0.883}  & \textbf{0.962} & \textbf{0.982} \\
\cline{2-12}
 & 2023 & Lee~\etal~\cite{Scaleaware_visualInertial_selfsup_scaled} & 1 &  M+Pose &  0.141 &1.117 &5.435 & 0.223 & 0.804  &0.942 &0.977 \\
\cline{2-12}
 & 2024\textdagger & Kinoshita~\& Nishino~\cite{CameraHeightDoesnChange} & 1 &  M+SI &  \underline{0.108} &\underline{0.785} &4.736 & 0.195 & 0.871  &0.958 &\underline{0.981} \\
\cline{2-12}
&  \multicolumn{2}{c|}{MonoPP (ours) } & 1 &  M+camH &   \textbf{0.107}  &   0.835 &   \underline{4.658}  &   \textbf{0.186}  &   \textbf{0.891}  &  \textbf{0.962}  &   \textbf{0.982}   \\   
\hline

\end{tabular*}
\caption{Comparison of our method to existing self-supervised approaches on the KITTI~\cite{geiger2012kitti} Eigen split~\cite{eigen2015predicting}. There are two separated tables, the upper one is dedicated for the comparison of scale-invariant depth, which means the predicted depth still needs to be scaled, hence all methods still need to calculate the scale from the ground-truth. The lower table focuses on comparing against the methods that predicts scaled depth. The best results in each subsection are in \textbf{bold} second best are \underline{underlined}. As shown, Our method outperforms other methods in predicting scaled metric depth estimation. All comparison is done for the medium resolution (640 x 192). \textbf{M} stands for training by monocular videos, and \textbf{S} includes stereo data as well. \textbf{SC*} stands for predicting a scale consistent output, which may still need GT for scaling. \textbf{Pose} for utilizing the pose information, \textbf{V} for utilizing the vehicle's velocity, and \textbf{camH} for utilizing initial camera height from the ground, and \textbf{SI} for scraping large-dataset from the internet while training.$\uparrow$ higher values are better. $\downarrow$ lower values are better. \textdagger~ This is an arxiv pre-print which first published in 2023, but these are their new results reported in 2024. ** is a baseline that we implemented to predict post-processed metric-scaled depth from Monodepth2, scaled by the GT camera height, as illustrated in~\cref{sec:scale_recov}}

\label{tab:kitti_tab_appendix}
\end{table*}
\end{document}